\documentclass{gammarobot}
\usepackage[utf8]{inputenc}
\usepackage{amsmath,amssymb,amsfonts,mathtools}
\usepackage{amsmath,amsfonts,bm}

\def\eqref#1{equation~\ref{#1}}

\def\1{\bm{1}}

\DeclareMathAlphabet{\mathsfit}{\encodingdefault}{\sfdefault}{m}{sl}
\SetMathAlphabet{\mathsfit}{bold}{\encodingdefault}{\sfdefault}{bx}{n}

\usepackage{epigraph,wrapfig,fancyhdr}

\fancypagestyle{gammaTeleAI}{
  \fancyhf{}
  \fancyfoot[C]{\thepage}

}
\makeatletter
\let\ps@plain\ps@gammaTeleAI
\makeatother

\title{ECHO: In-Context Behavior foundation model for Humanoid Whole-Body Control}
\author[1,2,*]{Xingyi Wang}
\author[2,3,*]{Bin Li}
\author[2]{Jiyuan Shi}
\author[2]{Chenyun Zhang}
\author[2,4]{Weiji Xie}
\author[2,4]{Xiangyu Miao}
\author[2,4]{Junsong Wu}
\author[1]{Wei Song}
\author[1,\dagger]{Shiqiang Zhu}
\author[2,5,\dagger]{Chenjia Bai}
\affiliation[1]{Zhejiang University}
\affiliation[2]{Institute of Artificial Intelligence (TeleAI), China Telecom}
\affiliation[3]{ShanghaiTech University}
\affiliation[4]{Shanghai Jiao Tong University}
\affiliation[5]{Gamma Robotics}
\contribution[*]{Equal contribution (co-first authors)}
\contribution[\dagger]{Corresponding authors}
\date{\today}
\project{\url{https://mwondering.github.io/echo-bfm/}}
\correspondence{Chenjia Bai (\email{baicj@chinatelecom.cn})}

\abstract{
Humanoid behavior foundation models (BFMs) acquire diverse skills
through large-scale motion imitation and massively parallel
reinforcement learning, yet broad behavioral capability does not
ensure precise execution across physical conditions.
Without distinguishing the underlying dynamics, policies may
favor conservative compromises over environment-specific precision.
We introduce \textbf{ECHO}, an in-context BFM that turns interaction
experience into environment-specific control, separating general
behavioral capability from dynamics-specific execution.
Mixed and Annealed Sampling first establishes a behavior
prior in a single pretraining stage by balancing targeted practice
with broad motion coverage.
Building on this prior, Context Representation and Residual
Adaptation learns structured dynamics representations from
interaction histories to guide residual corrections to the frozen
prior, enabling adaptation without test-time parameter updates.
Extensive experiments show that ECHO outperforms all
baseline methods across diverse motions and physical conditions.
Moreover, deployment on the Unitree G1 demonstrates whole-body teleoperation
spanning extreme acrobatics and terrain interaction, alongside
payload adaptation in manipulation and whole-body control. Page is
available at \href{https://mwondering.github.io/echo-bfm/}{https://mwondering.github.io/echo-bfm/}.
}

\begin{document}
\maketitle

\begin{figure}[htbp]
    \centering
    \includegraphics[width=1.0\linewidth]{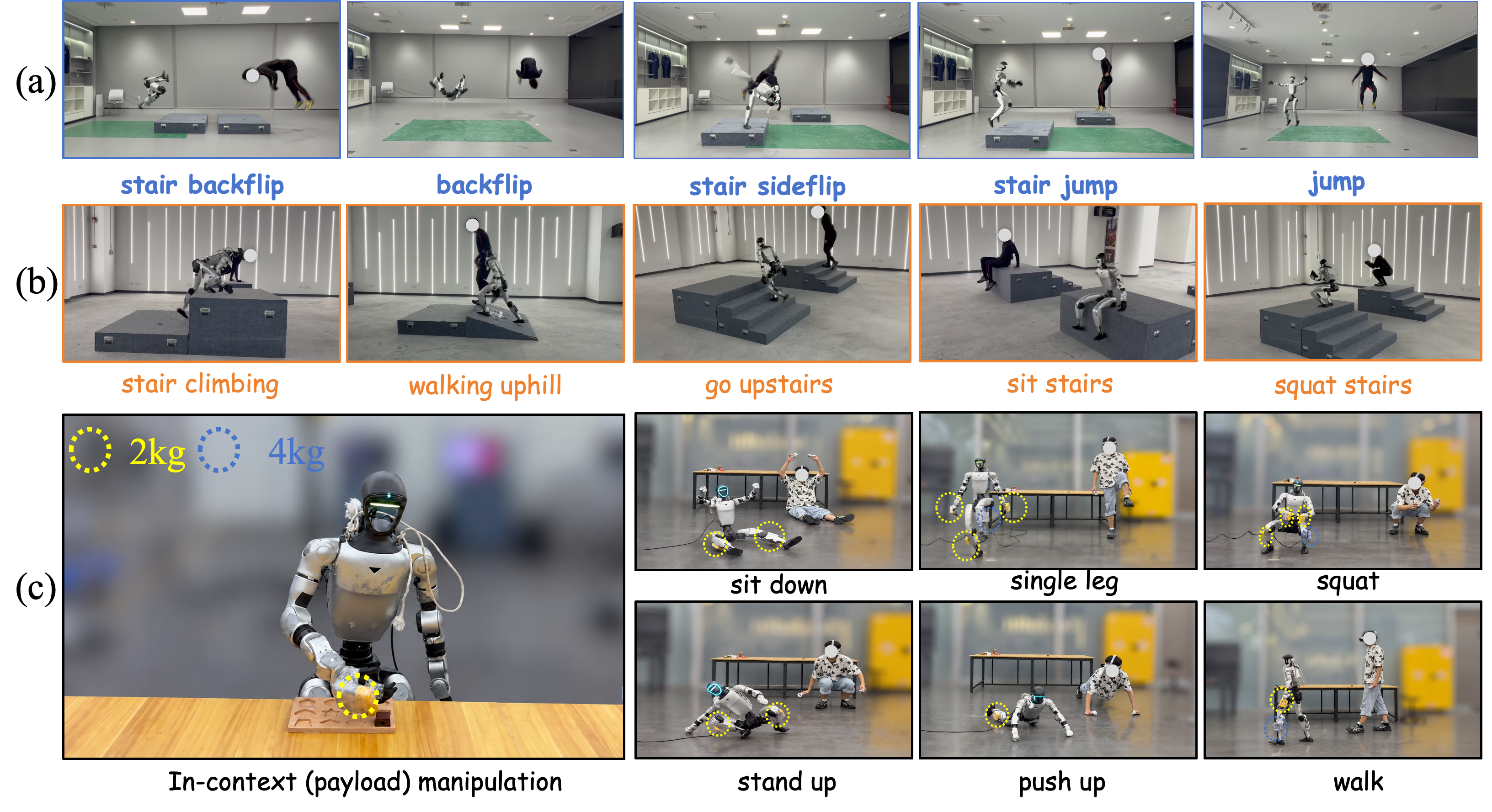}
    \caption{\textbf{ECHO: Broad whole-body capabilities with in-context adaptation.}
A shared behavior prior supports whole-body teleoperation spanning
extreme acrobatics~(a) and terrain-interactive behaviors~(b),
while interaction-derived context enables adaptation to
payload-induced dynamics changes during manipulation and
whole-body control~(c).}
    \label{fig:teaser}
\end{figure}

\section{Introduction}

\epigraph{
\textit{For the things we have to learn before we can do them, we learn by doing them.}
}{
Aristotle, \textit{Nicomachean Ethics}
}

Humanoid robots are expected to execute diverse whole-body behaviors accurately, responsively, and robustly in human-centric environments. Large-scale motion imitation and massively parallel reinforcement
learning (RL) have advanced this goal through behavior foundation
models (BFMs)~\citep{luo2026sonic,zeng2025behavior,qi2026humanoid,zeng2026scaling,mimiclite2026,liu2026heft,cheng2026gigabrain,ma2026extreme,li2026bfm,chen2026holomotion,liu2026humantracker}.
Yet a broad behavior repertoire alone does not ensure precise
execution across physical conditions: the same reference motion
may require different control responses as payload, friction,
or actuator dynamics vary. Without adequately distinguishing these dynamics, a policy trained
across environments may learn conservative compromises that favor
broad reliability over environment-specific tracking
precision~\citep{mehta2020active,wang2026halo}.
The key challenge is therefore to enable a BFM to understand the underlying dynamics across environments
and adapt to different physical conditions.

Recent work in robotic manipulation has demonstrated in-context
adaptation through human demonstrations~\citep{shah2025mimicdroid}
or retrieved interaction experience~\citep{chen2026zeva,zhang2025omnivic},
without updating the model weights at test time.
These advances motivate equipping humanoid BFMs with the ability
to learn environmental dynamics from interaction context and
tailor whole-body actions accordingly.
As a BFM executes diverse behaviors, its actions and the resulting
state transitions provide evidence about how the underlying dynamics
shape its response.
By encoding this evidence into structured dynamics representations,
the model can learn control-relevant properties of the current
environment and use them to guide subsequent actions.
Interaction histories thus serve not merely as records of past
execution, but as context for environment-specific action corrections.
This allows a shared behavior prior to pursue the same
objective with control responses tailored to different physical
conditions, adapting its execution without test-time parameter updates.

To this end, we introduce \textbf{ECHO}, an in-context behavior
foundation model for humanoid whole-body control that separates
general behavioral capability from dynamics-specific execution. Specifically,
we first train a motion-tracking prior, \textbf{ECHO-PRE}, using
Mixed and Annealed Sampling (MIAS).
By balancing targeted practice with broad motion coverage,
MIAS enables a single policy to jointly learn everyday motions,
highly dynamic skills~\citep{wang2026omnixtreme}, contact-rich floor behaviors, and terrain
motions within one pretraining stage
(Table~\ref{tab:echo_capabilities}).
we introduce Context Representation and Residual Adaptation
(CRA), which learns structured representations of the underlying
environmental dynamics from the prior's interaction histories.
CRA combines multi-step dynamics prediction with
physical neighborhood soft alignment to capture control-relevant
physical information across behaviors.
A context-conditioned residual policy then uses these representations
to generate action corrections tailored to the learned dynamics,
adapting the frozen prior's execution.

We evaluate ECHO from four fundamental perspectives:
(1) \emph{tracking fidelity}, covering
challenging high-dynamic skills (Figure~\ref{fig:teaser} (a)) and terrain motion (Figure~\ref{fig:teaser} (b));
(2) \emph{in-context adaptation}, assessing the benefit of
interaction context under nominal and heavily randomized dynamics;
(3) \emph{representation quality}, examining whether
representations distinguish physical environments across motions; and
(4) \emph{real-world applicability}, demonstrating whole-body teleoperation and payload adaptation
on the Unitree G1 (Figure~\ref{fig:teaser} (c)).
Extensive experimental results show that MIAS strengthens the general behavior
prior, while CRA learns physically informative context that
supports more precise execution across varying dynamics without
test-time parameter updates.

Our contributions are summarized as follows:
\begin{itemize}
\item We introduce \textbf{ECHO}, an in-context behavior foundation
model that separates general behavioral competence from
environment-specific execution, enabling a single whole-body
behavior prior to retain broad capabilities while adapting to varying physical dynamics.

\item We develop two complementary components:
\textbf{Mixed and Annealed Sampling (MIAS)} for acquiring a broad
whole-body repertoire in a single pretraining stage, and
\textbf{Context Representation and Residual Adaptation (CRA)} for
turning interaction histories into physically grounded context
that guides dynamics-specific corrections to the frozen prior.

\item Extensive experiments show that ECHO improves tracking under
substantial dynamics variation, learns context representations that
remain physically informative across motions, and transfers to
real-world whole-body teleoperation and payload adaptation.
\end{itemize}

\begin{table}[t]
  \centering
  \caption{\textbf{Comparison of whole-body control capabilities.} ECHO-PRE covers all four capability categories within a single pretrained policy. $\checkmark$ indicates a demonstrated capability, while $\times$ indicates one not reported in the corresponding work.}
  \begingroup
  \small
  \setlength{\tabcolsep}{5pt}
  \renewcommand{\arraystretch}{1.15}
  \newcommand{\yes}{\ensuremath{\checkmark}}
  \newcommand{\no}{\ensuremath{\times}}

  \resizebox{\linewidth}{!}{\begin{tabular}{@{}lcccccc@{}}
      \noalign{\hrule height 0.8pt}
      & \shortstack{Extreme play}
      & \shortstack{Extreme teleop}
      & \shortstack{Ground interaction}
      & \shortstack{Terrain interaction} \\
      \hline
      SONIC~\citep{luo2026sonic}
          & \yes &\no & \yes & \no     \\
      GigaBrain-WBC-0.5~\citep{cheng2026gigabrain}
          & \yes &\no & \no & \yes      \\
      HEFT~\citep{liu2026heft}
          & \yes &\no  & \no  & \no   \\
      Extreme-RGMT~\citep{ma2026extreme}
          & \yes &\yes  & \no  & \no    \\
      \hline
      \textbf{ECHO-PRE}
          & \yes &\yes & \yes & \yes   \\
      \noalign{\hrule height 0.8pt}
  \end{tabular}}
  \endgroup
  \label{tab:echo_capabilities}
\end{table}
\section{Related Work}

\paragraph{Behavior Foundation Models.}
Behavior foundation models learn reusable whole-body control priors.
BFM~\cite{zeng2025behavior} supports diverse control interfaces
through generative behavioral representations, while
BFM-Zero~\cite{li2026bfm} uses unsupervised RL to unify motion-,
goal-, and reward-conditioned control in a shared latent space.
SONIC~\cite{luo2026sonic}, Scaling BFM~\cite{zeng2026scaling},
Humanoid-GPT~\cite{qi2026humanoid}, and
HoloMotion-1~\cite{chen2026holomotion} explore complementary
data, model, and training scaling strategies to improve tracking
and generalization, while Mimiclite~\cite{mimiclite2026}
emphasizes efficient generalized motion tracking.
Beyond tracking, related systems support reactive motion
planning~\citep{chen2026reactivebfm}, terrain-aware
execution~\citep{wang2026perceptive}, and environmental
interaction~\citep{cheng2026gigabrain}.
HEFT~\cite{liu2026heft} and Extreme-RGMT~\cite{ma2026extreme}
further extend capabilities to heavy-payload teleoperation and
highly dynamic skills, respectively.
Our ECHO complements these advances by combining broad skill
acquisition via MIAS with
dynamics-specific residual adaptation through CRA.

\paragraph{In-Context Learning and Dynamics Adaptation.}
Algorithm Distillation~\cite{laskin2022context} and
Prompt-DT~\cite{xu2022prompting} demonstrate adaptation through
learning histories and demonstration prompts, respectively,
without test-time parameter updates.
In robotics, earlier work combines universal policies with online system identification to adapt control to unknown dynamics~\citep{yu2017preparing}. RMA~\cite{kumar2021rma} and dual-history
policies~\cite{li2025reinforcement} exploit proprioceptive
histories for dynamics adaptation, while
DWL~\cite{gu2024advancing} uses denoising reconstruction for
implicit system identification.
LocoFormer~\cite{liu2025locoformer} leverages cross-trial
Transformer memory to adapt locomotion to unseen morphologies
and dynamics.
For humanoid control, Any2Track~\cite{zhang2025track} combines
world-model-based dynamics representations with layer-wise
adapters in a frozen tracker, while HoRD~\cite{wang2026hord}
integrates history-conditioned RL with online distillation. FADA~\cite{xie2026fada} separates planning from inverse dynamics
and adapts the latter using target-domain rollouts. In contrast,
ECHO combines multi-step dynamics prediction with cross-motion
physical neighborhood soft alignment to learn structured context
representations, then trains an RL-based residual policy~\citep{johannink2019residual,wang2026general} conditioned on the learned in-context representation
with both the encoder and behavior prior frozen.

\begin{figure}[htbp]
    \centering
    \includegraphics[width=1.0\linewidth]{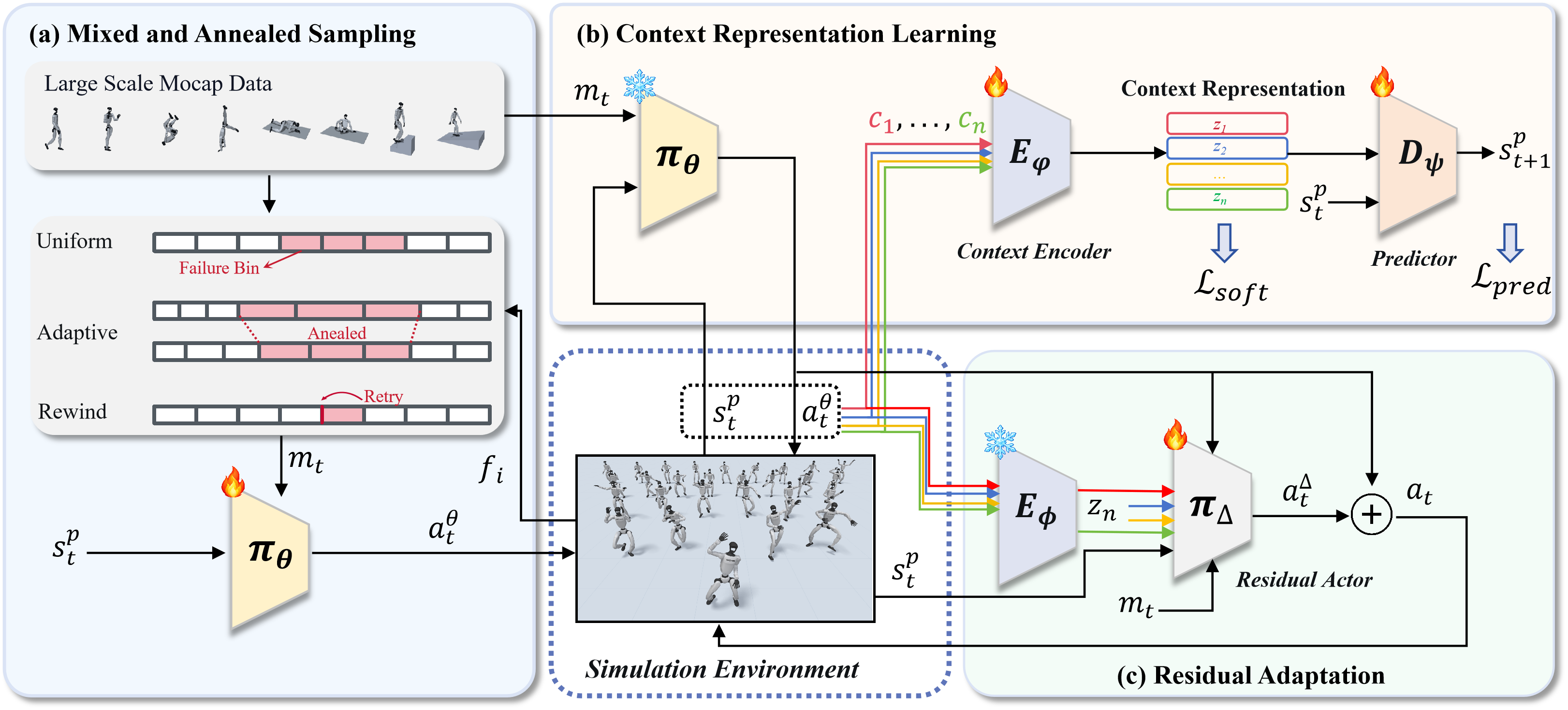}
    \caption{\textbf{Method Overview.} \textbf{ECHO-PRE $\pi_{\theta}$} is pretrained using \textbf{Mixed and Annealed Sampling},
which balances targeted practice with broad motion coverage.
\textbf{Context Representation Learning} extracts in-context
latents from the frozen prior's interaction context histories through
dynamic prediction and physical neighborhood soft alignment.
\textbf{Residual Adaptation} learns in-context
corrections with both the pretrained prior and context encoder frozen, enabling
in-context adaptation without test-time updates. Here, $s_t^p$ denotes the proprioceptive state, $m_t$ the reference
motion, $a_t^{\theta}$ the action produced by $\pi_{\theta}$, $f_i$ the failure rate,
$c_i$ the physical configuration of the $i$-th environment,
$E_{\varphi}$ the context encoder, $z_i$ the context representation
of $c_i$, $D_{\psi}$ the dynamics predictor, $\pi_{\Delta}$ the
in-context residual actor, $a_t^{\Delta}$ the residual action produced
by $\pi_{\Delta}$, and $a_t$ the final applied action.}
    \label{fig:pipeline}
\end{figure}

\section{Method}
  \label{sec:method}
As illustrated in Figure~\ref{fig:pipeline}, we first formulate In-Context BFM under unobserved environmental dynamics
(Sec.~\ref{sec:problem_formulation}). We then introduce MIAS to
pretrain ECHO-PRE, balancing broad motion coverage with targeted
practice on difficult segments (Sec.~\ref{sec:behavior_prior}).
Building on this frozen prior, CRA enables effective in-context
adaptation by inferring environmental dynamics from interaction
history and translating the resulting context into action
corrections tailored to the current environment
(Sec.~\ref{sec:in_context_tracking}).

\subsection{Problem Formulation}
\label{sec:problem_formulation}

We formulate humanoid control under varying physical conditions as a
family of Markov decision processes (MDPs) indexed by an environmental
configuration $c \in \mathcal{C}$:
\begin{equation}
    \mathcal{M}_c
    =
    \langle
    \mathcal{S}, \mathcal{A}, \mathcal{T}_c, \mathcal{R}, \gamma
    \rangle,
    \label{eq:context_mdp}
\end{equation}
where $c$ specifies properties such as mass distribution,
friction, and payload, and $\mathcal{T}_c$ is the corresponding
transition dynamics. The configuration $c$ is not directly observable at
deployment.

Following prior work~\citep{zeng2026scaling}, we use humanoid whole-body
motion tracking as the proxy task for BFM pretraining. A conventional
domain-randomized BFM $\pi(a_t \mid s_t^p, m_t)$ generates actions $a_t$ from proprioceptive states $s_t^p$ and reference-motion
goals $m_t$ using a policy shared across heterogeneous dynamics,
which can compromise environment-specific performance. To this end, we introduce \textbf{In-Context BFM}, which learns the underlying environmental dynamics from interaction context $\mathcal{H}_t$ and enables adaptive control under varying environmental configurations $c$.

\textbf{We formally define the interaction context for humanoid control as the history
of proprioceptive states and actions}:
\begin{equation}
    \mathcal{H}_t
    \triangleq
    \left(
    s_0^p, a_0, s_1^p, a_1, \ldots, s_{t-1}^p, a_{t-1}, s_t^p
    \right).
    \label{eq:interaction_context}
\end{equation}
This history captures the humanoid's responses to past actions and
thus provides evidence about the environmental configuration $c$
governing its dynamics. We learn a context encoder $E_\varphi$ to extract a latent representation
of the environmental configuration $c$ from interaction history
$\mathcal{H}_t$:
\begin{equation}
    z_t
    =
    E_\varphi(\mathcal{H}_t)
    \approx
    z_c
    \triangleq
    g(c),
    \label{eq:environment_representation}
\end{equation}
Here, $g:\mathcal{C}\rightarrow\mathcal{Z}$ maps each environmental
configuration to a dynamics-relevant latent representation $z_c$,
and $z_t$ denotes its context-dependent estimate.
The resulting In-Context BFM  acts according to
$\pi(a_t \mid s_t^p, m_t, z_t)$, enabling adaptation without
direct access to privileged parameters.

With the learned encoder $E_\varphi$ held fixed, our objective
is to maximize the expected discounted return over a distribution
$p(c)$ of environmental configurations:
\begin{equation}
    \max_{\pi}\;
    \mathbb{E}_{\substack{
        c \sim p(c) \\
        \tau \sim p_\pi(\tau \mid c)
    }}
    \left[
        \sum_{t=0}^{T-1} \gamma^t r_t
    \right],
    \label{eq:bfm_objective}
\end{equation}
where $r_t$ denotes the motion-tracking reward and
$p_\pi(\tau \mid c)$ is the distribution of length-$T$ rollouts
induced by the In-Context BFM $\pi$ under $\mathcal{T}_c$.

\subsection{Learning a General Behavior Prior via MIAS}
\label{sec:behavior_prior}
In-context adaptation across diverse whole-body behaviors requires a
sufficiently general behavior prior. We first train the behavior prior $\pi_\theta$, termed
\textbf{ECHO-PRE}, in a single stage with MIAS, using motion
tracking as the proxy task.
MIAS combines failure rewind, uniform exploration, and failure-adaptive
sampling to balance broad motion coverage with targeted practice on
difficult segments, and gradually reduces sampling concentration to
consolidate the learned repertoire.

\paragraph{Adaptive weight.}
We partition reference motions into temporal bins and estimate
their difficulty using empirical failure rates $f_i$~\citep{luo2026sonic,liao2026beyondmimic}.
We prioritize bins with observed failures through a length-weighted,
temperature-scaled distribution:
\begin{equation}
\widetilde p_i
\propto
\ell_i \exp(f_i/T)\,\mathbf 1\{f_i>0\},
\label{eq:adaptive_prioritization}
\end{equation}
where $\ell_i$ is the number of valid frames in bin $i$ and
$T>0$ controls sampling sharpness.
To prevent excessive concentration on individual segments or motions,
we cap and redistribute probability mass at both levels, yielding
the adaptive distribution $p^{\mathrm A}$.
Counting rules, the no-failure fallback, and probability caps
are detailed in Appendix~\ref{app:adaptive_sampling}.

\paragraph{Mixed and annealed sampling.}
Difficult segments warrant targeted practice without indefinitely
restricting motion coverage. MIAS therefore mixes three
sampling branches and progressively relaxes the concentration
of the adaptive branch.

\emph{Mixed sampling.}
Let $\mathcal{P}_{\mathrm R}$, $\mathcal{P}_{\mathrm U}$, and
$\mathcal{P}_{\mathrm A}$ denote the distributions over bins induced by rewind, uniform, and adaptive sampling, respectively.
We sample bins from the failure-conditioned mixture:
\begin{equation}
\mathcal{P}_d
=
d\rho_{\mathrm R}\mathcal{P}_{\mathrm R}
+
(1-d\rho_{\mathrm R})
\bigl[
\rho_{\mathrm U}\mathcal{P}_{\mathrm U}
+
(1-\rho_{\mathrm U})\mathcal{P}_{\mathrm A}
\bigr],
\label{eq:termination_conditioned_sampling}
\end{equation}
where $d\in\{0,1\}$ indicates whether the preceding trajectory ended
in a tracking failure, and $\rho_{\mathrm R},\rho_{\mathrm U}\in[0,1]$
specify the rewind probability conditional on failure and the uniform
probability conditional on not rewinding, respectively.
Rewind restarts the failed motion at the beginning of its failure bin.
Uniform and adaptive sampling select a bin uniformly from $\mathcal I$
or according to $p^{\mathrm A}$, respectively, and then sample a valid
frame uniformly within that bin.

\emph{Annealed concentration.}
To gradually broaden practice, we increase the sampling temperature $T$
and decrease the probability cap $C$ using a shared annealing function:
\begin{equation}
\theta_k
=
\mathcal A\!\left(
k;\theta_{\mathrm{start}},\theta_{\mathrm{end}},K_\theta
\right),
\qquad \theta\in\{T,C\},
\label{eq:sampling_annealing}
\end{equation}
where $k$ is the training iterations and $\mathcal A$ performs
cosine interpolation with independent horizons $K_T$ and $K_C$,
as specified in Appendix~\ref{app:adaptive_sampling}.

\subsection{In-Context Adaptation via CRA}
\label{sec:in_context_tracking}
CRA first learns a context encoder from the prior's
rollouts using dynamics prediction and physically grounded
representation objectives, then freezes the context encoder and trains
a residual policy to translate context into action corrections.

\paragraph{Interaction context.}
Let $s_t^p$ denote noisy proprioception and $a_t$ the applied action.
At decision time $t$, the context $\mathcal H_t$
contains only completed, valid transitions preceding that decision.
We instantiate the context with short- and long-term histories:
\begin{equation}
\begin{aligned}
\mathcal H_t
=(\mathcal S_t,\mathcal M_t),
\qquad
z_t=E_\varphi(\mathcal H_t),
\end{aligned}
\label{eq:echo_context}
\end{equation}
The short-term history $\mathcal S_t$ preserves fine-grained
recent responses, while the long-term memory $\mathcal M_t$
provides complementary dynamics evidence from earlier interactions
across motions.
Complete chunks are retained in $\mathcal M_t$ across
reference-motion changes and episode resets within the same
physical session, whereas $\mathcal S_t$ is cleared at these
boundaries.
The physical context remains fixed within a session, and a new
session clears all memory.

\paragraph{Predictive in-context dynamics.}
To encourage the context latent to encode control-relevant dynamics,
we jointly train $E_\varphi$ and a forward predictor $D_\psi$ on rollouts
collected by the frozen tracker.
Let $y_t$ contain the privileged robot state, foot kinematics,
contact forces, and binary contact indicators, and let $a_t$
denote the applied action.
Conditioned on the context $z_t$, the predictor estimates future
states over a horizon of $K_{\mathrm p}$ steps:
\begin{equation}
\hat y_{t+k}
=
D_\psi\!\left(\mathcal H_{t+k-1},a_{t+k-1};z_t\right),
\label{eq:echo_prediction}
\end{equation}
where $\mathcal H_{t+k-1}$ denotes the predictor's rolling input
history, initialized from the observed trajectory and updated
with recorded states or model predictions during training.
The context $z_t$ is held fixed throughout the prediction horizon,
and all predictions use the recorded applied target sequence.
We minimize the multi-step prediction objective:
\begin{equation}
\mathcal L_{\mathrm{pred}}
=
\mathbb E_{\substack{
\tau\sim\mathcal D\\
\mathbf H\sim\nu(\cdot\mid\tau)
}}\!\left[
\frac{1}{K_{\mathrm p}}
\sum_{k=1}^{K_{\mathrm p}}
\ell_{\mathrm{pred}}(\hat y_{t+k},y_{t+k})
\right],
\label{eq:echo_prediction_loss}
\end{equation}
where $\mathcal D$ is the rollout-segment distribution and
$\nu(\cdot\mid\tau)$ is the weighted distribution over rolling-history
sequences $\mathbf H=(\mathcal H_t,\ldots,\mathcal H_{t+K_{\mathrm p}-1})$.

\paragraph{Physically grounded context geometry.}
We complement predictive supervision with
\emph{physical neighborhood soft alignment} to preserve
neighborhoods among physical conditions.
Specifically, we align cross-motion context neighborhoods $P$
with physical-parameter neighborhoods $Q$:
\begin{equation}
\mathcal L_{\mathrm{soft}}
=
\frac{1}{n}\sum_{i=1}^{n}
D_{\mathrm{KL}}\!\left(Q_{i:}\,\|\,P_{i:}\right),
\label{eq:echo_soft_loss}
\end{equation}
where $n$ is the number of eligible randomized systems.
More details are in
Appendix~\ref{app:context_geometry}.

Then, we jointly optimize the context encoder and forward predictor with
\begin{equation}
\mathcal L_{\mathrm{ctx}}
=
\mathcal L_{\mathrm{pred}}
+
\mathcal L_{\mathrm{soft}}.
\label{eq:echo_context_loss}
\end{equation}
Privileged states and physical parameters are used only during
training.

\paragraph{In-context residual adaptation.}
The learned context representation provides evidence about
environmental dynamics, which the residual policy translates
into environment-specific action corrections.
We freeze both the context encoder $E_\varphi$ and
ECHO-PRE $\pi_\theta$ and train a residual actor $\pi_\Delta$.
The predictor is not used in this stage or at deployment.
The final action combines the prior's command $a_t^{\theta}$ with a
context-conditioned correction $a_t^{\Delta}$:
\begin{equation}
a_t^\theta=\pi_\theta(s_t^p),
\qquad
a_t^\Delta=\pi_\Delta(s_t^p,m_t, z_t,a_t^\theta),
\qquad
a_t=a_t^\theta+a_t^\Delta,
\label{eq:echo_control}
\end{equation}
where $a_t$ is the final applied action.
Each completed valid transition updates the interaction history
through Eq.~\eqref{eq:echo_context}, closing the
interaction--context--control loop.

\section{Experiments}
\subsection{Set up}
We use the Unitree G1 for both training and evaluation. We evaluate ECHO extensively in simulation, including tracking fidelity (Section~\ref{sec:exp_motion_tracking_benchmark}), in-context adaptation (Section~\ref{sec:exp_context_using}), and in-context representation analysis (Section~\ref{sec:exp_latent_structure}). We further validate ECHO on the real robot in Section~\ref{sec:exp_real_world_experiments}.

To enable fair comparison with existing BFMs while evaluating in-context ability under broader dynamics variations, we design two physical randomization settings. \textbf{Necessary Domain Randomization (NDR)} contains the minimum set of DR required for sim-to-real transfer and is designed to remain close to the settings commonly adopted by existing baselines. We pretrain ECHO-PRE only under NDR to acquire the general behavior prior.

We further introduce \textbf{Heavy Domain Randomization (HDR)}, which extends NDR with substantially larger dynamics variations, including additional payloads on the wrists and lower legs. Starting from the same frozen ECHO-PRE backbone, we train CRA under NDR and HDR, resulting in \textbf{ECHO-N} and \textbf{ECHO-H}, respectively. The detailed randomization ranges of Nominal environment, NDR and HDR are summarized in Table~\ref{tab:ndr_hdr}.

We train ECHO using PPO~\citep{schulman2017proximal} in Mjlab ~\citep{zakka2026mjlab} for behavior pretraining and in-context adaptation, leveraging more than 1,200 hours of filtered high-quality motion data. The composition of the training dataset is illustrated in Figure~\ref{fig:dataset}.

\subsection{Evaluation of Motion Tracking Fidelity}
\label{sec:exp_motion_tracking_benchmark}
We evaluate the contributions of MIAS and CRA to motion tracking.
To assess MIAS, we compare \textbf{ECHO-PRE} with
\textbf{HEFT}~\citep{liu2026heft}, \textbf{SONIC v1.1}~\citep{luo2026sonic},
and \textbf{ECHO-PRE w/o MIAS}, which uses uniform motion sampling.
To assess CRA, we compare \textbf{ECHO-N} against its pretrained
\textbf{ECHO-PRE} backbone.
Our evaluation covers two complementary benchmarks:
\textbf{AMASS-2000-Random} tests generalization to everyday motions
from AMASS, which is excluded from ECHO's training data, while
\textbf{BONES-SEED-100-HighDynamic} tests challenging motions within
the training distribution. All evaluations use nominal dynamics
without domain randomization. We report success rate (SR) and
motion-tracking errors, with dataset construction and metric
definitions detailed in Appendix~\ref{app:motion_tracking_evaluation}.

As shown in Table~\ref{tab:motion_tracking_fidelity}, MIAS improves
tracking on both benchmarks relative to uniform sampling, and
ECHO-PRE compares favorably with the external baselines.
CRA further reduces tracking errors,
showing that interaction context improves execution precision
even under nominal dynamics.

\begin{table}[t]
\centering
\caption{\textbf{Simulation evaluation of BFMs' motion tracking fidelity.}
MIAS improves tracking performance, while CRA further enhances fidelity across both everyday and high-dynamic motions. Results are reported as mean $\pm$ sample standard deviation across three evaluation runs with independent action-noise seeds and noise standard deviation $\sigma=0.01$. Bold and underlined values indicate the best and second-best results within each dataset, respectively.}
\label{tab:motion_tracking_fidelity}

\begingroup
\footnotesize
\setlength{\tabcolsep}{3pt}
\renewcommand{\arraystretch}{1.18}
\newcommand{\result}[2]{#1{\scriptsize\,\(\pm\)\,#2}}

\begin{tabular*}{\linewidth}{@{\extracolsep{\fill}}lccccc@{}}
  \hline
  Method
  & SR $\uparrow$
  & $E_{\mathrm{g\text{-}MPJPE}}\downarrow$
  & $E_{\mathrm{MPJPE}}\downarrow$
  & $E_{\mathrm{acc}}\downarrow$
  & $E_{\mathrm{vel}}\downarrow$ \\
  \hline

  \multicolumn{6}{c}{\textbf{AMASS-2000-Random}} \\
  \hline

  SONIC v1.1
  & \result{0.9633}{0.0003}
  & \result{316.651}{1.649}
  & \result{30.411}{0.015}
  & \result{1.135}{0.002}
  & \result{3.770}{0.006} \\

  HEFT
  & \result{0.9785}{0.0009}
  & \result{261.573}{2.352}
  & \result{28.524}{0.009}
  & \result{1.062}{0.001}
  & \result{3.103}{0.004} \\

  \hline
  ECHO-PRE w/o MIAS
  & \result{0.9720}{0.0005}
  & \result{331.235}{2.132}
  & \result{32.580}{0.018}
  & \result{1.243}{0.003}
  & \result{4.170}{0.008} \\
  ECHO-PRE
  & \result{\textbf{0.9832}}{0.0012}
  & \result{\underline{207.995}}{0.970}
  & \result{\underline{26.595}}{0.007}
  & \result{\underline{0.932}}{0.003}
  & \result{\underline{3.036}}{0.006} \\

  \textbf{ECHO-N}
  & \result{\underline{0.9827}}{0.0003}
  & \result{\textbf{203.149}}{0.289}
  & \result{\textbf{25.577}}{0.024}
  & \result{\textbf{0.917}}{0.003}
  & \result{\textbf{2.956}}{0.007} \\

  \hline
  \multicolumn{6}{c}{\textbf{BONES-SEED-100-HighDynamic}} \\
  \hline

  SONIC v1.1
  & \result{0.5800}{0.0100}
  & \result{343.124}{4.421}
  & \result{61.791}{2.615}
  & \result{4.086}{0.031}
  & \result{10.406}{0.102} \\

  HEFT
  & \result{0.6200}{0.0173}
  & \result{391.094}{30.758}
  & \result{59.562}{0.135}
  & \result{3.232}{0.017}
  & \result{8.229}{0.138} \\

  \hline
  ECHO-PRE w/o MIAS
  & \result{{0.4500}}{0.0100}
  & \result{{365.075}}{5.270}
  & \result{{64.580}}{3.025}
  & \result{{4.297}}{0.042}
  & \result{{11.712}}{0.145} \\
  ECHO-PRE
  & \result{\underline{0.7200}}{0.0000}
  & \result{\underline{244.772}}{3.143}
  & \result{\underline{56.521}}{0.335}
  & \result{\underline{2.801}}{0.012}
  & \result{\underline{7.507}}{0.034} \\

  \textbf{ECHO-N}
  & \result{\textbf{0.7633}}{0.0115}
  & \result{\textbf{235.410}}{4.140}
  & \result{\textbf{55.392}}{0.415}
  & \result{\textbf{2.766}}{0.005}
  & \result{\textbf{7.360}}{0.017} \\

  \hline
\end{tabular*}
\endgroup
\end{table}
\subsection{Evaluating the ability of using context}
\label{sec:exp_context_using}
In this section, we compare ECHO with representative adaptation methods for legged robot control. We consider three baselines. \textbf{RMA}~\citep{kumar2021rma} learns a privileged teacher conditioned on ground-truth dynamics parameters and subsequently distills its adaptation capability into a student policy; we report both the \textbf{RMA-Teacher} and \textbf{RMA-Student} for comparison. \textbf{Any2Track}~\cite{zhang2025track} learns a latent dynamics representation from interaction history and regularizes the representation with a world-model prediction objective. We further design a \textbf{Vanilla} baseline, which uses the same residual-policy architecture but receives only the observations available to ECHO-PRE, without any latent dynamics representation.

For a fair comparison, all methods share the same frozen ECHO-PRE backbone and residual-policy architecture, and are trained under the same HDR setting. We evaluate all methods on AMASS-2000-Random under both the nominal environment and the HDR environment. The evaluation metrics are identical to those used in Section~\ref{sec:exp_motion_tracking_benchmark}.

As shown in Table ~\ref{tab:context_utilization}, ECHO-H achieves performance comparable to RMA-Teacher in the nominal environment, without requiring privileged dynamics parameters at inference.
Under HDR, ECHO-H achieves the best results on four of the five metrics, improving SR by 0.95 percentage points and reducing global MPJPE by 3.0\% relative to RMA-Teacher. ECHO-H also outperforms RMA-Student across all HDR metrics and consistently improves over Vanilla in both environments. In contrast, the Any2Track baseline provides no consistent gains over Vanilla, suggesting that introducing a history-derived latent alone does not guarantee effective adaptation in this setting. Together, these results suggest that ECHO learns control-relevant representations of environmental dynamics that the residual policy can effectively exploit to improve tracking under substantial dynamics variation.

  \begin{table}[t]
  \centering
  \caption{\textbf{Simulation evaluation of BFMs' in-context ability.} ECHO-H effectively exploits interaction context to improve tracking under heavily randomized dynamics (HDR).
Simulation evaluation of BFMs' in-context ability. Bold and underlined values indicate the best and second-best results within each environment, respectively.}
  \label{tab:context_utilization}

  \begingroup
  \footnotesize
  \setlength{\tabcolsep}{3pt}
  \renewcommand{\arraystretch}{1.18}
  \newcommand{\result}[2]{#1{\scriptsize\,\(\pm\)\,#2}}

  \begin{tabular*}{\linewidth}{@{\extracolsep{\fill}}lccccc@{}}
    \hline
    Method
    & SR $\uparrow$
    & $E_{\mathrm{g\text{-}MPJPE}}\downarrow$
    & $E_{\mathrm{MPJPE}}\downarrow$
    & $E_{\mathrm{acc}}\downarrow$
    & $E_{\mathrm{vel}}\downarrow$ \\
    \hline

    \multicolumn{6}{c}{\textbf{Nominal Environment}} \\
    \hline

    Vanilla
    & \result{0.8892}{0.0003}
    & \result{194.541}{1.585}
    & \result{27.286}{0.016}
    & \result{0.881}{0.008}
    & \result{2.887}{0.006} \\

    Any2Track
    & \result{0.8743}{0.0008}
    & \result{193.623}{0.738}
    & \result{27.846}{0.002}
    & \result{0.873}{0.008}
    & \result{2.883}{0.005} \\

    RMA-Teacher
    & \result{\textbf{0.9798}}{0.0003}
    & \result{\textbf{189.908}}{0.220}
    & \result{\underline{24.062}}{0.010}
    & \result{\underline{0.849}}{0.007}
    & \result{\underline{2.772}}{0.004} \\

    RMA-Student
    & \result{\underline{0.9763}}{0.0008}
    & \result{190.329}{0.590}
    & \result{\textbf{24.026}}{0.007}
    & \result{0.851}{0.008}
    & \result{2.774}{0.005} \\

    \hline

    \textbf{ECHO-H}
    & \result{0.9748}{0.0010}
    & \result{\underline{190.088}}{0.718}
    & \result{24.496}{0.014}
    & \result{\textbf{0.846}}{0.008}
    & \result{\textbf{2.760}}{0.006} \\

    \hline
    \multicolumn{6}{c}{\textbf{HDR Environment}} \\
    \hline

    Vanilla
    & \result{0.8632}{0.0008}
    & \result{230.001}{0.833}
    & \result{31.541}{0.018}
    & \result{1.088}{0.003}
    & \result{3.701}{0.004} \\

    Any2Track
    & \result{0.8455}{0.0009}
    & \result{227.549}{0.999}
    & \result{31.939}{0.012}
    & \result{1.087}{0.003}
    & \result{3.709}{0.003} \\

    RMA-Teacher
    & \result{\underline{0.8827}}{0.0021}
    & \result{\underline{216.826}}{1.462}
    & \result{\textbf{30.063}}{0.017}
    & \result{\underline{1.052}}{0.002}
    & \result{\underline{3.591}}{0.002} \\

    RMA-Student
    & \result{0.8700}{0.0009}
    & \result{224.568}{0.894}
    & \result{30.542}{0.014}
    & \result{1.065}{0.002}
    & \result{3.648}{0.001} \\

    \hline

    \textbf{ECHO-H}
    & \result{\textbf{0.8922}}{0.0016}
    & \result{\textbf{210.351}}{1.017}
    & \result{\underline{30.463}}{0.008}
    & \result{\textbf{1.039}}{0.003}
    & \result{\textbf{3.530}}{0.003} \\

    \hline
  \end{tabular*}
  \endgroup
  \end{table}
\subsection{Analyzing the learned context representations}
\label{sec:exp_latent_structure}

We examine whether CRA learns environment representations that remain
consistent across motions while retaining physically meaningful
information. We collect trajectories by rolling out ECHO-PRE across
40 LAFAN motions in 1,024 HDR environments and evaluate all variants
on 325,193 shared valid history windows. We evaluate environment identification, cross-motion retrieval, separation, and linear readout of payload mass and friction. See Appendix~\ref{app:context_representation_evaluation} for details.

\begin{figure}[t]
  \centering
  \begin{minipage}[t]{0.43\linewidth}
    \centering
    \includegraphics[width=\linewidth]{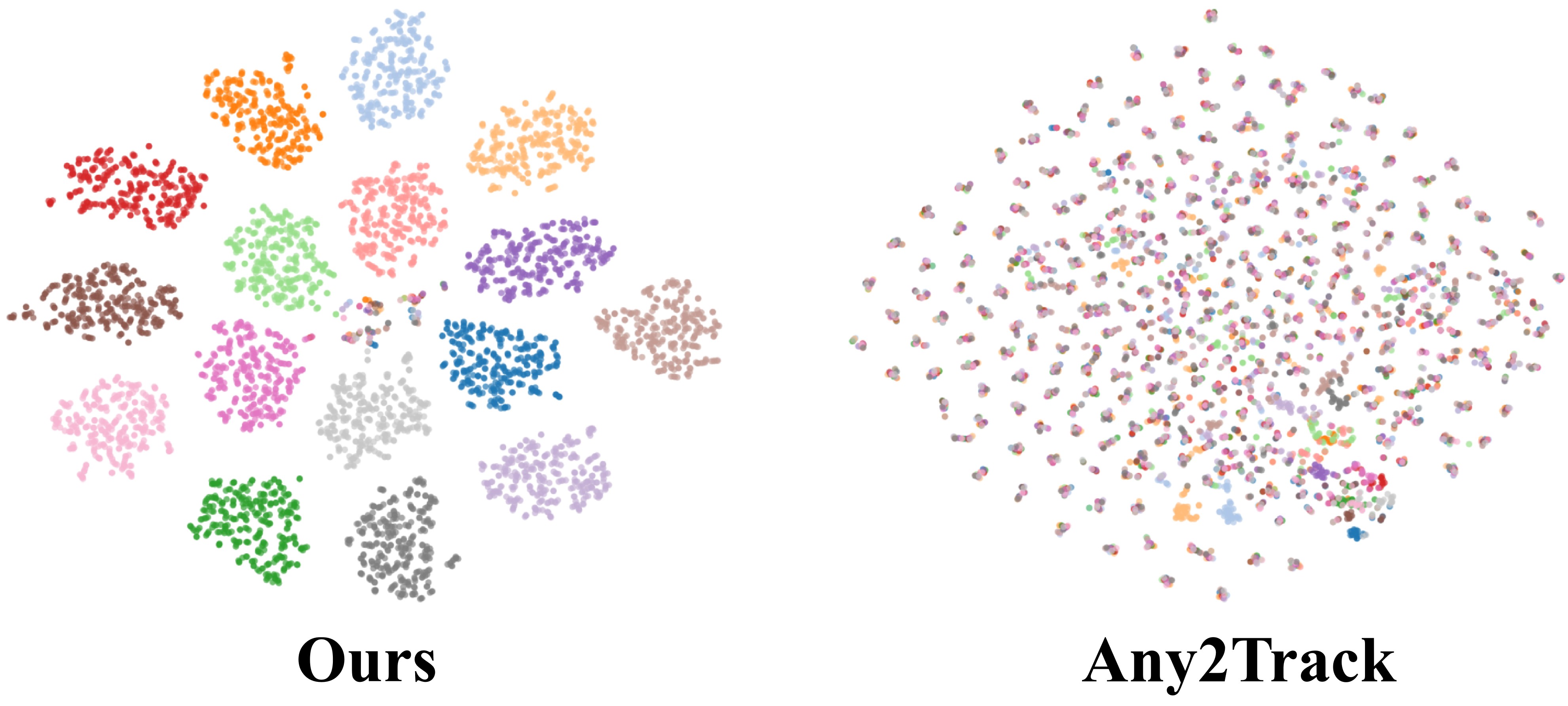}
    \caption{\textbf{Context representation comparison.} t-SNE visualizations of ECHO (left) and Any2Track (right), with colors indicating environments. ECHO shows clearer environment-wise grouping across motions.}
    \label{fig:context_latent_structure}
  \end{minipage}
  \hfill
  \begin{minipage}[t]{0.55\linewidth}
    \centering
    \includegraphics[width=\linewidth]{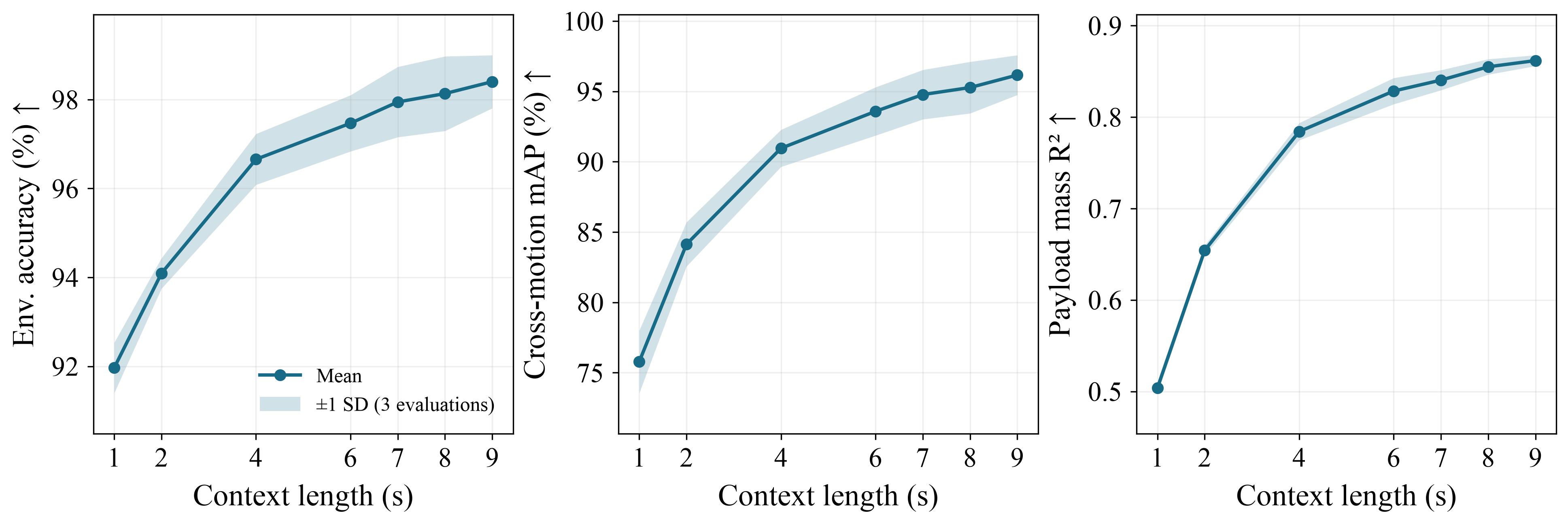}
   \caption{\textbf{Context-length scaling of environment representations.}
Linear environment identification accuracy, cross-motion retrieval mAP,
and payload-mass linear readout $R^2$ on LAFAN-40 (left to right).
The points and shaded
bands denote the mean and sample standard deviation over three evaluations.}
    \label{fig:context_scaling}
  \end{minipage}
\end{figure}

As reported in Table~\ref{tab:context_representation}, ECHO achieves
97.478\% mean environment identification accuracy and 93.872\% mean
cross-motion retrieval mAP, with mean linear-readout $R^2$ scores of
0.8357 for payload mass and 0.9208 for friction. These results indicate that the learned
representations capture environment information consistently across
motions while keeping physical properties linearly accessible.
The t-SNE visualization~\citep{maaten2008visualizing} in
Fig.~\ref{fig:context_latent_structure} provides qualitative support,
showing clearer environment-wise grouping than our Any2Track
implementation~\citep{zhang2025track} on 16 randomly selected
environments. Additional visualizations on AMASS are provided
in Appendix~Fig.~\ref{fig:amass100_tsne_128env}.

The loss ablations reveal complementary roles of soft alignment
and the predictor objective. Removing physical-neighborhood soft
alignment reduces cross-motion mAP from 93.872\% to 91.282\% and the
silhouette score from 0.4642 to 0.4337, suggesting weaker cross-motion
consistency and environment separation. In contrast, removing the
predictor leaves retrieval and separation scores nearly unchanged,
but reduces payload mass $R^2$ from 0.8357 to 0.4777 and friction $R^2$
from 0.9208 to $-0.0132$. This contrast highlights the predictor's
contribution to making physical properties linearly accessible,
beyond merely distinguishing environments.

\paragraph{Scaling interaction context.}
Our context encoder supports longer interaction histories by aggregating
additional long-term chunks into a fixed-size memory representation.
We fix the short-term history at 1 second (50 steps) and compare a
short-only baseline with total context budgets of 2, 4, 6, 7, 8, and 9 seconds.
Each variant is trained separately for 5,000 updates using the prediction
loss and soft-alignment loss. Evaluation uses shared, unaveraged history
windows from 40 LAFAN~\citep{harvey2020robust} motions, and we report means
over three evaluations. Figure~\ref{fig:context_scaling} shows consistent
improvements with longer context: from 50 to 450 steps (1 to 9 seconds),
mean linear environment identification accuracy increases from 91.967\%
to 98.403\%, and cross-motion retrieval mAP from 75.778\% to 96.165\%.
Payload mass $R^2$ increases from 0.5037 to 0.8618, showing improved linear
readability of physical properties. These results demonstrate that our
method can scale its interaction context to learn more informative
environment representations with better cross-motion consistency.
Complete results and protocol details are provided in
Appendix~\ref{app:context_representation_evaluation}
(Table~\ref{tab:rep_context_scaling}).

  \begin{table}[t]
  \centering
  \caption{\textbf{Evaluation of context representations on LAFAN-40.}
  Results are mean $\pm$ sample standard deviation over five evaluation seeds with independently sampled window starts and observation noise. Encoder checkpoints, the 1,024 physical environments, and probe splits are fixed; all methods share trajectories and valid windows within each seed. Mass denotes added payloads. Bold and underlined indicate the best and second-best results, ranked before rounding.}
  \label{tab:context_representation}
  \begingroup
  \footnotesize
  \setlength{\tabcolsep}{3pt}
  \renewcommand{\arraystretch}{1.18}
  \newcommand{\result}[2]{#1{\scriptsize\,\(\pm\)\,#2}}

  \begin{tabular*}{\linewidth}{@{\extracolsep{\fill}}lccccc@{}}
    \hline
    Method
    & \shortstack{Env. Acc.\\(\%) $\uparrow$}
    & \shortstack{mAP\\(\%) $\uparrow$}
    & Silhouette $\uparrow$
    & Mass $R^2\uparrow$
    & Friction $R^2\uparrow$ \\
    \hline

    w/o soft alignment
    & \result{97.125}{0.636}
    & \result{91.282}{1.233}
    & \result{0.4337}{0.0084}
    & \result{\underline{0.7869}}{0.0061}
    & \result{\underline{0.9069}}{0.0078} \\

    w/o predictor
    & \result{\underline{97.141}}{0.743}
    & \result{\underline{93.761}}{1.482}
    & \result{\underline{0.4555}}{0.0143}
    & \result{0.4777}{0.0025}
    & \result{-0.0132}{0.0003} \\

    \hline
    \textbf{ECHO}
    & \result{\textbf{97.478}}{0.762}
    & \result{\textbf{93.872}}{1.450}
    & \result{\textbf{0.4642}}{0.0105}
    & \result{\textbf{0.8357}}{0.0080}
    & \result{\textbf{0.9208}}{0.0088} \\
    \hline
  \end{tabular*}
  \endgroup
  \end{table}

\subsection{Real-world experiments}
\label{sec:exp_real_world_experiments}

We demonstrate ECHO's diverse whole-body behaviors through real-time
teleoperation, including crawling, highly dynamic motions, and terrain
traversal (Figure~\ref{fig:teaser}, Figure~\ref{fig:echo-pre-show}, and
the accompanying video).
As shown in Figure~\ref{fig:echo-in-context-show}, in-context adaptation
enables ECHO-H to learn context-dependent control under HDR and maintain
precise tracking with added payloads. Even with 10\,kg of payloads
distributed across its hands and legs (three 2\,kg loads and one 4\,kg
load), ECHO-H can still stand on one leg, squat, and bend forward. For quantitative evaluation, we execute four reference motions and
compute MPJPE-L from joint positions recorded during deployment.
We compare ECHO-N with ECHO-PRE, HEFT, and SONIC v1.1 without additional
payloads, and ECHO-H with Vanilla and RMA-Student under a 2\,kg load on
each lower leg. As reported in Table~\ref{tab:real_mpjpe}, ECHO-N and
ECHO-H achieve the lowest mean tracking errors across all four motions in
their respective settings, supporting the effectiveness of
context-conditioned adaptation for real-world motion tracking.

\section{Conclusion}
\label{sec:conclusion}

We presented ECHO, an in-context behavior foundation model that separates general behavioral capability from dynamics-specific execution. MIAS acquires a broad whole-body repertoire, while CRA learns physical dynamics from interaction context to guide residual corrections to the frozen behavior prior. Experiments and real-world deployment on the Unitree G1 demonstrate improved tracking, whole-body teleoperation, and adaptation to added payloads. These results show that interaction context enables a shared behavior prior to adapt to its physical environment, combining behavioral breadth with dynamics-specific precision without test-time parameter updates.

\subsection*{AI use statement}

In this work, we used generative AI tools, to
assist with translation and to obtain feedback on research
methodology, experimental design, and the interpretation of
experimental results. We did not use generative AI tools to
generate synthetic datasets, develop theoretical models or conceptual frameworks,
or formulate mathematical claims. Additionally, we used these
tools for manuscript drafting
and organization, language editing, and LaTeX formatting.
The authors reviewed and revised all AI-assisted content, checked
references and related-work descriptions against the original
publications, and verified technical descriptions and result
interpretations against the implementation and experimental
records. We take full responsibility for the final content of
this work, including text, claims, and artifacts produced with
the assistance of generative AI.

\FloatBarrier
\bibliographystyle{plainnat}
\bibliography{references}

\clearpage
\beginappendix

\section{Details of Methods}

\paragraph{Adaptive sampling.}
\label{app:adaptive_sampling}

Let $\mathcal I$ denote the set of valid temporal bins, where
$i=(m,b)$ identifies bin $b$ of motion $m$.
For each bin, we maintain completed-visit and failure counts,
$V_i$ and $F_i$, and estimate its failure rate as $f_i=F_i/V_i$
for $V_i>0$.
A visit is completed when a trajectory leaves the bin or terminates.
We compute the preliminary sampling distribution as
\begin{equation}
\begin{aligned}
s_i
&=
\ell_i
\exp\!\left(\frac{f_i-f_{\max}}{T}\right)
\mathbf 1\{f_i>0\},\\
\widetilde p_i
&=
\begin{cases}
\displaystyle
\frac{s_i}{\sum_{j\in\mathcal I}s_j},
& \displaystyle \sum_{j\in\mathcal I}s_j>0,\\[8pt]
\displaystyle
\frac{\ell_i}{\sum_{j\in\mathcal I}\ell_j},
& \text{otherwise},
\end{cases}
\end{aligned}
\label{eq:adaptive_score}
\end{equation}
where $f_{\max}=\max_{j\in\mathcal I}f_j$ is subtracted for
numerical stability.
Length weighting accounts for partially filled bins and provides
a length-proportional fallback when no failures have been observed.
We then cap and redistribute probability mass to obtain
$p^{\mathrm A}$ satisfying
\begin{equation}
\sum_{i\in\mathcal I}p_i^{\mathrm A}=1,
\qquad
0\leq p_i^{\mathrm A}\leq\frac{C}{N},
\qquad
\sum_{i\in\mathcal I_m}p_i^{\mathrm A}\leq\frac{C}{M}
\quad \forall m,
\label{eq:sampling_caps}
\end{equation}
where $N=|\mathcal I|$, $M$ is the number of motions, and
$\mathcal I_m$ contains the bins of motion $m$.
The multiplier $C\geq1$ is chosen so that the constraints are
jointly feasible, and redistribution preserves relative weights
wherever capacity permits.

\paragraph{Mixed sampling.}
In practice, we set $\rho_{\mathrm R}=1/3$ and $\rho_{\mathrm U}=1/2$ in
Eq.~\eqref{eq:termination_conditioned_sampling}, yielding
\begin{equation}
\mathcal{P}_d
=
\frac{d}{3}\mathcal{P}_{\mathrm R}
+
\frac{3-d}{6}
\bigl(
\mathcal{P}_{\mathrm U}+\mathcal{P}_{\mathrm A}
\bigr).
\label{eq:app_branch_selection}
\end{equation}
After a tracking failure ($d=1$), all three components receive
equal weight $1/3$.
After a non-failure timeout or normal motion completion ($d=0$),
rewind is disabled, and uniform and adaptive sampling each
receive weight $1/2$.
A tracking failure coinciding with a timeout is assigned $d=1$.

The rewind component $\mathcal{P}_{\mathrm R}$ selects the failure
bin of the preceding motion and restarts at the beginning of
that bin.
For each valid bin $i\in\mathcal I$, the uniform and adaptive
components satisfy
$\mathcal{P}_{\mathrm U}(i)=1/|\mathcal I|$ and
$\mathcal{P}_{\mathrm A}(i)=p_i^{\mathrm A}$, respectively.
Both components then sample a valid frame $\tau$ uniformly within
the selected bin.
Uniform sampling initializes at $\tau$, whereas adaptive sampling
additionally applies a random backward offset to include the
lead-up to difficult segments:
\begin{equation}
t_0=\max\{0,\tau-u\},
\qquad
u\sim\operatorname{Uniform}\{0,\ldots,W-1\},
\label{eq:app_adaptive_backward_offset}
\end{equation}
where $\tau$ and $t_0$ are frame indices within the selected motion,
and $W$ is a positive integer specifying the backward-offset
window in frames.

\paragraph{Annealing schedule.}
The annealing function in Eq.~\eqref{eq:sampling_annealing}
interpolates from an initial value $a$ to a final value $b$
over a horizon of $K$ iterations:
\begin{equation}
\mathcal A(k;a,b,K)
=
a+\frac{b-a}{2}
\left[
1-\cos\!\left(\pi\min\{k/K,1\}\right)
\right],
\label{eq:app_cosine_annealing}
\end{equation}
where $k\geq0$ and $K>0$.
For the sampling temperature, we use
$(a,b,K)=(T_{\mathrm{start}},T_{\mathrm{end}},K_T)$,
with $T_{\mathrm{end}}>T_{\mathrm{start}}>0$.
For the cap multiplier, we use
$(a,b,K)=(C_{\mathrm{start}},C_{\mathrm{end}},K_C)$,
with $C_{\mathrm{start}}>C_{\mathrm{end}}\geq1$.
Each schedule reaches its final value at the corresponding horizon
and remains constant thereafter.
All scheduled cap values are chosen to keep the constraints in
Eq.~\eqref{eq:sampling_caps} jointly feasible.

\paragraph{Prediction loss.} The per-step prediction loss is
\begin{equation}
\ell_{\mathrm{pred}}(\hat y,y)
=
\sum_{g\in\mathcal G}\alpha_g
\operatorname{Huber}\!\left(e_g(\hat y,y)\right)
+
\alpha_{\mathrm{bin}}
\operatorname{BCE}(\hat y^{\mathrm{bin}},y^{\mathrm{bin}}).
\label{eq:echo_prediction_components}
\end{equation}
Here, $\mathcal G$ comprises root position and orientation,
linear and angular velocity, joint position and velocity,
foot kinematics, and contact force.
Each $e_g$ is a scaled component-wise error, with rotation-vector
errors for orientation and Huber losses~\citep{huber1992robust} averaged over coordinates.
The binary cross entropy~\citep{bengio2017deep} term supervises contact occurrence.

\paragraph{Physical neighborhood soft alignment.}
\label{app:context_geometry}
For $n$ eligible randomized systems in a minibatch, let $\xi_i$
denote the range-normalized, group-balanced physical-parameter
vector of system $i$, with $d_{ij}=\|\xi_i-\xi_j\|_2$.
Let $\bar z_i$ and $\bar z_i'$ denote the mean context embeddings
from its current and archived histories, respectively.
A \emph{current history} contains the completed
state--action--next-state transitions available at the corresponding
training sample's time; ``current'' is relative to that sample
and does not imply recently collected data.
An \emph{archived history} is an earlier, previously stored
interaction window from the same environment instance and the
same physical session, during which the physical parameters
remain unchanged.
The paired current and archived windows correspond to different
reference motions and do not overlap in time.
Only systems with an eligible history pair are included in
the neighborhood loss. Our neighborhood-distribution matching is related to probabilistic
knowledge transfer~\citep{passalis2018learning}, but uses
physical-parameter distances to supervise cross-motion
context similarities.
We construct the physical and latent neighborhood distributions as
\begin{equation}
\begin{aligned}
Q_{ij}
&=
\operatorname{softmax}_{j}\!\left(
-(\log 2)\left(\frac{d_{ij}}{h_c}\right)^2
\right),\\
P_{ij}
&=
\operatorname{softmax}_{j}\!\left(
\frac{\bar z_i^\top\bar z_j'}{\tau_z}
\right),
\end{aligned}
\label{eq:echo_neighborhood}
\end{equation}
where softmax normalizes each row over all $n$ systems,
$h_c>0$ controls the physical-neighborhood bandwidth, and
$\tau_z>0$ is the latent similarity temperature.
The neighborhood loss in Eq.~\eqref{eq:echo_soft_loss}
is the average row-wise KL divergence:
\begin{equation}
\mathcal L_{\mathrm{soft}}
=
\frac{1}{n}
\sum_{i=1}^{n}
\sum_{j=1}^{n}
Q_{ij}\log\frac{Q_{ij}}{P_{ij}}.
\label{eq:app_soft_loss}
\end{equation}
Matching neighborhoods across different motions encourages the
learned relationships to reflect physical conditions rather than
motion identity.

\paragraph{Joint optimization.}
We jointly optimize the context encoder $E_\varphi$ and forward
predictor $D_\psi$ using the total objective in
Eq.~\eqref{eq:echo_context_loss}.
Each loss term is averaged over its eligible samples.
Privileged states and physical parameters are used only during
training; the context encoder receives interaction history alone.

\section{Details of training}
\begin{figure}[htbp]
    \centering
    \includegraphics[width=1.0\linewidth]{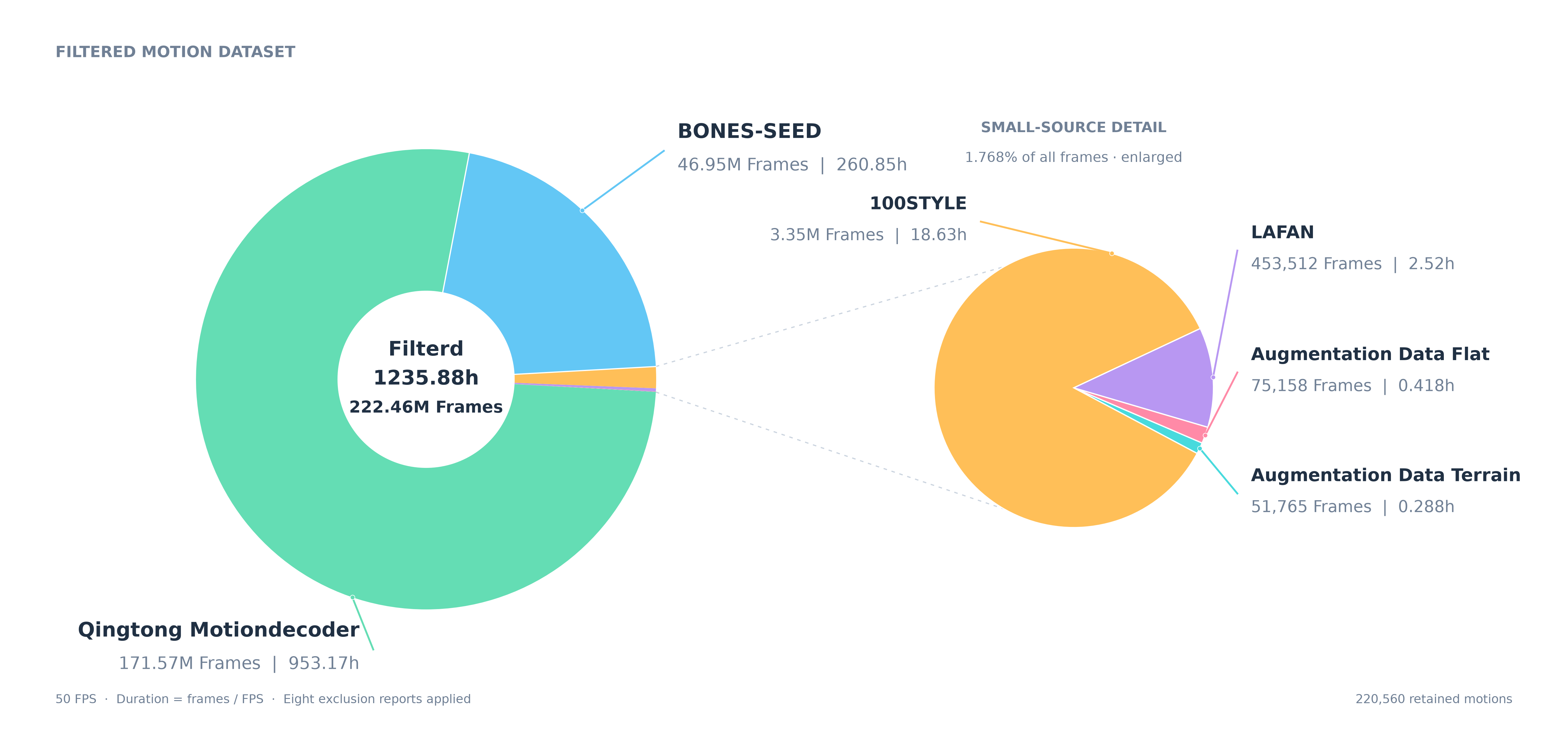}
    \caption{\textbf{Composition of the filtered training corpus.}
The corpus comprises 1,235.88 hours of motion data
(222.46 million frames).
The left chart shows source proportions by frame count,
while the right chart provides an enlarged breakdown of
the smaller sources.}
    \label{fig:dataset}
\end{figure}

\paragraph{Datasets.}
As summarized in Figure~\ref{fig:dataset}, our filtered training
corpus contains 220,560 motion sequences and approximately
222.46 million frames, corresponding to 1,235.88 hours at
50 frames per second.
It comprises MotionDecode~\citep{chingmu2026motiondecode}
(953.17 hours), BONES-SEED~\citep{studio142bones}
(260.85 hours), 100STYLE~\citep{mason2022real}
(18.63 hours), and LAFAN1~\citep{harvey2020robust}
(2.52 hours).
The corpus additionally includes two augmentation subsets,
\emph{Augmentation Data Flat} and
\emph{Augmentation Data Terrain}, containing 75,158 and
51,765 frames, respectively.
All reported quantities refer to the data retained after filtering,
rather than the complete source datasets.
AMASS~\cite{mahmood2019amass} is excluded from training and
used to evaluate generalization to unseen motions.
Tables~\ref{tab:training_rewards} and~\ref{tab:training_observations}
summarize the rewards and observations used to train ECHO-PRE.
Tracking terms use exponential error shaping as in HEFT~\citep{liu2026heft}.
\paragraph{Reward and Observation Specifications}
\label{app:reward_observation}
Tables~\ref{tab:training_rewards} and~\ref{tab:training_observations}
summarize the rewards and observations used to train ECHO-PRE.
Tracking terms use exponential error shaping as in HEFT~\citep{liu2026heft}.

\paragraph{Domain randomization settings.}
Table~\ref{tab:ndr_hdr} summarizes the nominal configuration and
the two domain randomization settings, NDR and HDR.
The nominal environment uses the baseline physical parameters
with a foot friction coefficient of $0.6$, no additional payloads,
and no external force perturbations.
NDR follows the original tracker configuration, randomizing
torso mass and center of mass (COM), foot friction, controller
gains, and joint armature, while applying intermittent external
forces.
HDR retains all NDR ranges and additionally introduces four-limb
payload randomization, with $[0,\,2.5]$\,kg per hand and
$[0,\,4.0]$\,kg per shin (up to $13$\,kg in total), payload COM
offsets of $\pm 5$\,cm per axis, and inertia recomputed to account
for the added payloads.

\begin{table}[t]
\centering
\caption{\textbf{Reward terms used for ECHO-PRE training.}}
\label{tab:training_rewards}
\begingroup
\footnotesize
\rmfamily
\setlength{\tabcolsep}{3pt}
\renewcommand{\arraystretch}{1.18}
\begin{tabular*}{\linewidth}{@{\extracolsep{\fill}}lcl@{}}
\hline
\textbf{Reward term} & \textbf{Weight} $w_k$ & \textbf{Unweighted formula} $r_k$ \\
\hline
root\_pos\_tracking & 1.5 & $\exp[-\|p-\hat p\|_2/0.3]$ \\
root\_rot\_tracking & 1.0 & $\exp[-d(R,\hat R)/0.4]$ \\
root\_vel\_tracking & 1.0 & $\exp[-\|R^\top v-\hat R^\top\hat v\|_2/1.0]$ \\
root\_ang\_vel\_tracking & 1.0 & $\exp[-\|R^\top\omega-\hat R^\top\hat\omega\|_2/3.0]$ \\
keypoint\_pos\_tracking & 2.0 & $\exp[-\sum_i\|x_i^b-\hat x_i^b\|_2/(0.3K)]$ \\
keypoint\_vel\_tracking & 1.0 & $\exp[-\sum_i\|v_i^b-\hat v_i^b\|_2/(1.0K)]$ \\
keypoint\_rot\_tracking & 2.0 & $\exp[-\sum_i d(R_i^b,\hat R_i^b)/(0.4K)]$ \\
keypoint\_angvel\_tracking & 1.0 & $\exp[-\sum_i\|\omega_i^b-\hat\omega_i^b\|_2/(3.0K)]$ \\
joint\_pos\_tracking & 1.0 & $\exp[-\|q-\hat q\|_1/(0.5n)]$ \\
joint\_vel\_tracking & 0.5 & $\exp[-\|\dot q-\hat{\dot q}\|_1/(3.0n)]$ \\
survival & 3.0 & $1$ \\
joint\_vel\_l2 & $3.5\times10^{-4}$ & $-\sum_j\bar{\dot q}_j^{\,2}$ \\
waist\_joint\_acc\_l2 & $10^{-5}$ & $-\sum_{j\in\mathcal W}\langle\ddot q_j^2\rangle$ \\
joint\_power\_l1 & $2\times10^{-5}$ & $-\sum_j\langle|\tau_j\dot q_j|\rangle$ \\
knee\_joint\_power\_l1 & $5\times10^{-5}$ & $-\sum_{j\in\mathcal K}\langle|\tau_j\dot q_j|\rangle$ \\
action\_rate\_l2 & 0.3 & $-\|a_t-a_{t-1}\|_2^2$ \\
waist\_action\_rate\_l2 & $-5.0$ & $\sum_{j\in\mathcal W}(a_{t,j}-a_{t-1,j})^2$ \\
feet\_air\_time\_ref & 2.5 & $\sum_f c_{f,t}(1-c_{f,t-1})\min(T_{f,t}-0.25,0)$ \\
feet\_air\_time\_ref\_dense & 1.0 & $-\frac12\sum_f
\begin{cases}
1, & c_f\ne\hat c_f,\\
1-C_a(h_f^-), & c_f=\hat c_f=0,\\
C_c(h_f^+), & c_f=\hat c_f=1
\end{cases}$ \\
joint\_pos\_limits & 1.5 & $-10\sum_j\bigl([q_j^--q_j]_+ + [q_j-q_j^+]_+\bigr)$ \\
joint\_torque\_limits & 0.015 & $-\sum_j[|\tau_j|/(0.75\tau_j^{\max})-1]_+$ \\
\hline
\end{tabular*}
\par\vspace{0.4em}
\begin{minipage}{\linewidth}
\footnotesize
The per-step reward is $r_t=\Delta t\sum_k w_k r_k(t)$, with $\Delta t=0.02$\,s.
Hats denote reference states; $p,R,v,\omega$ denote the root state,
$q$ joint positions, $\tau$ joint torques, and $a$ raw actions.
$d$ is the relative rotation angle, $K=13$, $n=29$, and
$\mathcal W,\mathcal K$ are the waist and knee joint sets.
Keypoint quantities with superscript $b$ are root-relative and expressed
in the respective root frames. $\bar{\dot q}$ averages the last two physics
substeps; $\langle\cdot\rangle$ averages all four.
For the two feet, $c_f,\hat c_f$ are contact indicators and
$T_{f,t}=(1-c_{f,t-1})T_{f,t-1}+\Delta t(1-2|c_{f,t}-\hat c_{f,t}|)$,
initialized at zero.
$h_f^-,h_f^+$ are the minimum and maximum world-frame ankle/toe heights;
$C_a(h)=\operatorname{clip}_{[0,1]}((h-0.035)/0.12)$ and
$C_c(h)=\operatorname{clip}_{[0,1]}((h-0.035)/0.09)$.
Finally, $[x]_+=\max(x,0)$, $q_j^\pm$ are the 90\% soft position limits,
and $\tau_j^{\max}$ is the actuator torque limit.
\end{minipage}
\endgroup
\end{table}

\begin{table}[t]
\centering
\caption{\textbf{Observations used for ECHO-PRE training.}}
\label{tab:training_observations}
\begingroup
\footnotesize
\rmfamily
\setlength{\tabcolsep}{3pt}
\renewcommand{\arraystretch}{1.18}
\begin{tabular*}{\linewidth}{@{\extracolsep{\fill}}lp{0.65\linewidth}@{}}
\hline
\textbf{Signal} & \textbf{Definition} \\
\hline
\multicolumn{2}{l}{\textit{Actor observations}} \\
Proprioceptive history & Recent joint states, IMU signals, actions, and torques. \\
Estimated root height & Root height predicted by the state estimator. \\
Estimated foot contacts & Predicted foot contact probabilities. \\
Key-body state & Current key-body poses and velocities in the root frame. \\
Reference root/joint features & Root and joint targets derived from the denoised reference. \\
Reference key-body state & Key-body poses and velocities from the denoised reference. \\
State tracking errors & Errors in joint states, gravity, angular velocity, and height. \\
Key-body tracking errors & Key-body pose and velocity differences from the reference. \\
\hline
\multicolumn{2}{l}{\textit{Critic observations}} \\
Noisy reference features & Noisy motion targets and their tracking errors. \\
Noisy proprioceptive history & Joint states and IMU signals with observation noise. \\
Action history & Previous policy mean actions. \\
Privileged reference features & Clean reference root and joint motion targets. \\
Privileged key-body state & Key-body poses and velocities from the simulator. \\
Reference key-body pose/errors & Reference key-body poses and corresponding tracking errors. \\
Privileged proprioceptive history & True joint states, gravity, and root velocities. \\
Applied actions/torques & Applied joint position targets and actuator torques. \\
Foot contact state & Contact forces, states, durations, and reference contact flags. \\
\hline
\end{tabular*}
\endgroup
\end{table}

\begin{table}[t]
      \centering
      \caption{\textbf{Domain randomization ranges.}
      Nominal denotes the clean environment used for evaluation.
      NDR follows the original tracker configuration, while HDR extends
      NDR with four-limb payload randomization.}
      \label{tab:ndr_hdr}

      \begingroup
      \small
      \setlength{\tabcolsep}{3pt}
      \renewcommand{\arraystretch}{1.08}

      \begin{tabular*}{\linewidth}
          {@{\extracolsep{\fill}}lccc@{}}
          \hline
          Parameter
          & \textbf{Nominal}
          & NDR
          & HDR \\
          \hline

          Torso COM offset (cm/axis)
              & $\mathbf{0}$
              & $\pm 7.5$
              & $\pm 7.5$ \\

          Torso mass offset (kg)
              & $\mathbf{0}$
              & $[-1,\,1]$
              & $[-1,\,1]$ \\

          Foot friction coefficient
              & $\mathbf{0.6}$
              & $[0.3,\,2.0]$
              & $[0.3,\,2.0]$ \\

          Upper-limb $K_p$, $K_d$ scale
              & $\mathbf{1.0}$
              & $[0.9,\,1.1]$
              & $[0.9,\,1.1]$ \\

          Waist/lower-limb $K_p$, $K_d$ scale
              & $\mathbf{1.0}$
              & $[0.8,\,1.2]$
              & $[0.8,\,1.2]$ \\

          Joint armature scale
              & $\mathbf{1.0}$
              & $[0.8,\,1.2]$
              & $[0.8,\,1.2]$ \\

          External force (N/axis)
              & $\mathbf{0}$
              & $\pm 10$
              & $\pm 10$ \\

          Force duration (s)
              & --
              & $[0.3,\,0.5]$
              & $[0.3,\,0.5]$ \\

          Force interval (s)
              & --
              & $[3,\,6]$
              & $[3,\,6]$ \\

          \hline

          Payload per hand (kg)
              & $\mathbf{0}$
              & $0$
              & $[0,\,2.5]$ \\

          Payload per shin (kg)
              & $\mathbf{0}$
              & $0$
              & $[0,\,4.0]$ \\

          Payload COM offset (cm/axis)
              & --
              & --
              & $\pm 5$ \\

          Total payload mass (kg)
              & $\mathbf{0}$
              & $0$
              & $[0,\,13]$ \\

          Payload-induced inertia
              & Unchanged
              & Unchanged
              & Recomputed \\

          \hline
      \end{tabular*}

      \vspace{0.4em}

      \endgroup
  \end{table}

\section{EVALUATION OF MOTION TRACKING FIDELITY}
\label{app:motion_tracking_evaluation}

\paragraph{Benchmark construction.}
We construct two benchmarks to assess complementary aspects of
motion tracking. \textbf{AMASS-2000-Random} contains 2,000 randomly
sampled motions from AMASS~\citep{mahmood2019amass}. Since AMASS is
excluded from ECHO's training data, this benchmark evaluates
ECHO's out-of-distribution generalization to diverse everyday
behaviors. \textbf{BONES-SEED-100-HighDynamic} contains 100
highly dynamic motions sampled from
BONES-SEED~\citep{studio142bones}: 33 ground-interaction motions,
33 extreme motions, and 34 dance motions. These motions belong
to the training distribution and assess the policy's mastery
of challenging behaviors.

\paragraph{Evaluation metrics and simulator.}
Following~\cite{luo2023perpetual,luo2026sonic}, we report success
rate (SR) and four discrepancy metrics between executed and
reference trajectories: global MPJPE
($E_{\mathrm{g\text{-}MPJPE}}$, mm), root-relative MPJPE
($E_{\mathrm{MPJPE}}$, mm), acceleration error
($E_{\mathrm{acc}}$, mm/frame$^2$), and velocity error
($E_{\mathrm{vel}}$, mm/frame). Higher SR and lower tracking
errors indicate better performance. All policies are evaluated
using the CPU implementation of MuJoCo~\citep{todorov2012mujoco}
under nominal dynamics, without domain randomization during
evaluation.

\section{Context Representation Evaluation}
\label{app:context_representation_evaluation}

\paragraph{Sampling and splits.}
Each environment--motion pair contributes up to eight independent
sessions, with one latent extracted after 500 warm-up steps.
ECHO variants are evaluated after 5,000 training iterations;
the default encoder produces a 64-dimensional context from at most
350 completed interactions.
Following grouped evaluation~\citep{roberts2017crossvalidation},
we split motions into five folds, keeping each motion's windows together.
For classification, three folds fit the probe and one selects
hyperparameters; the probe is then refitted on all four non-test folds.
Retrieval uses these four folds as the gallery.
Physical readout additionally splits environments into five folds,
yielding 25 tests that exclude both test environments and test motions
from probe fitting and validation. Preprocessing uses training data
only. LAFAN motions belong to the encoder's pretraining corpus.

\paragraph{Metrics.}
We report frozen-encoder linear classification
accuracy~\citep{chen2020simclr}, full-gallery
mAP~\citep{manning2008ir}, and the Silhouette
score~\citep{rousseeuw1987silhouettes} using environment identities.
Retrieval treats windows from the same environment as positives;
both retrieval and Silhouette use Euclidean distances between
original latent vectors without additional normalization.
Ridge probes~\citep{hoerl1970ridge} predict payload mass and friction.
Probe scores pool out-of-fold predictions across valid windows;
mass $R^2$ averages the four limb-wise scores, while friction
$R^2$ evaluates a single coefficient.

\paragraph{Visualization.}
Both panels use the same windows and t-SNE settings:
perplexity 5, PCA initialization and 1,500 iterations.
The Any2Track comparison is qualitative because inputs, capacity,
history length, and training setup differ. To further evaluate out-of-distribution generalization, we visualize the learned context representations on the AMASS dataset using t-SNE.
\begin{table}[t]
\centering
\caption{\textbf{Context-length scaling on LAFAN-40.}
  Results are mean $\pm$ standard deviation over three evaluations.
  Bold and underlined values indicate the best and second-best results in each
  column, ranked before rounding. Arrows indicate the preferred direction.}
\label{tab:rep_context_scaling}
\begingroup
\footnotesize
\setlength{\tabcolsep}{3pt}
\renewcommand{\arraystretch}{1.18}
\providecommand{\result}[2]{#1{\scriptsize\,\(\pm\)\,#2}}
\begin{tabular*}{\linewidth}{@{\extracolsep{\fill}}lcccccc@{}}
\hline
Method & \shortstack{Linear Env.\\Acc. (\%)\\ $\uparrow$} & \shortstack{Cross-motion\\mAP (\%)\\ $\uparrow$} & \shortstack{Silhouette\\ $\uparrow$} & \shortstack{Payload\\mass $R^2$\\ $\uparrow$} & \shortstack{Payload\\CoM $R^2$\\ $\uparrow$} & \shortstack{Friction\\$R^2$\\ $\uparrow$} \\
\hline
Short only (50) & \result{91.967}{0.567} & \result{75.778}{2.225} & \result{0.2764}{0.0130} & \result{0.5037}{0.0054} & \result{0.0253}{0.0037} & \result{0.6899}{0.0169} \\
Context 100 & \result{94.092}{0.342} & \result{84.132}{1.571} & \result{0.3321}{0.0131} & \result{0.6542}{0.0055} & \result{0.0751}{0.0043} & \result{0.8007}{0.0123} \\
Context 200 & \result{96.657}{0.572} & \result{90.959}{1.318} & \result{0.4183}{0.0126} & \result{0.7842}{0.0094} & \result{0.0970}{0.0046} & \result{0.8834}{0.0059} \\
Context 300 & \result{97.468}{0.632} & \result{93.583}{1.716} & \result{0.4484}{0.0140} & \result{0.8284}{0.0143} & \result{\underline{0.1275}}{0.0069} & \result{0.9146}{0.0102} \\
Context 350  & \result{97.946}{0.791} & \result{94.776}{1.757} & \result{0.4693}{0.0136} & \result{0.8404}{0.0110} & \result{\textbf{0.1286}}{0.0058} & \result{0.9269}{0.0087} \\
Context 400 & \result{\underline{98.135}}{0.838} & \result{\underline{95.276}}{1.833} & \result{\underline{0.4768}}{0.0126} & \result{\underline{0.8550}}{0.0084} & \result{0.1263}{0.0065} & \result{\underline{0.9331}}{0.0086} \\
Context 450 & \result{\textbf{98.403}}{0.596} & \result{\textbf{96.165}}{1.414} & \result{\textbf{0.4964}}{0.0131} & \result{\textbf{0.8618}}{0.0059} & \result{0.1209}{0.0058} & \result{\textbf{0.9352}}{0.0071} \\
\hline
\end{tabular*}
\endgroup
\end{table}

\section{Real world experiments}

\begin{figure}
    \centering
    \includegraphics[width=\linewidth]{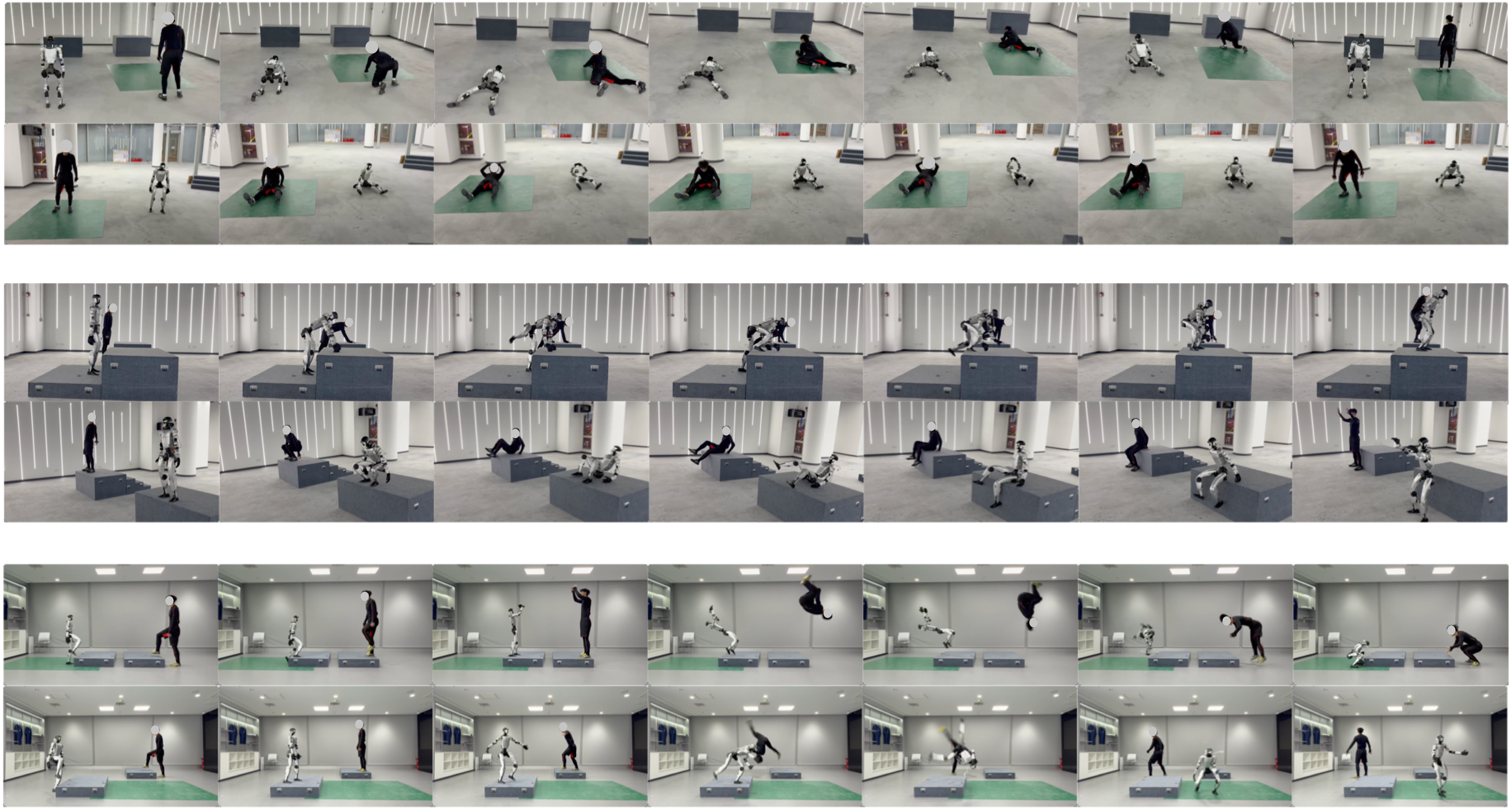}
    \caption{\textbf{Real-world whole-body teleoperation with ECHO-PRE.}
A single pretrained policy enables the Unitree G1 to track human
motions across ground interaction (top two rows), terrain interaction
(middle two rows), and extreme acrobatics (bottom two rows).
Examples include transitions between standing and floor-level
postures, platform climbing and descent, and aerial flips.
Each row shows successive frames of a teleoperated motion,
ordered from left to right.}
    \label{fig:echo-pre-show}
\end{figure}

\begin{figure}
    \centering
    \includegraphics[width=\linewidth]{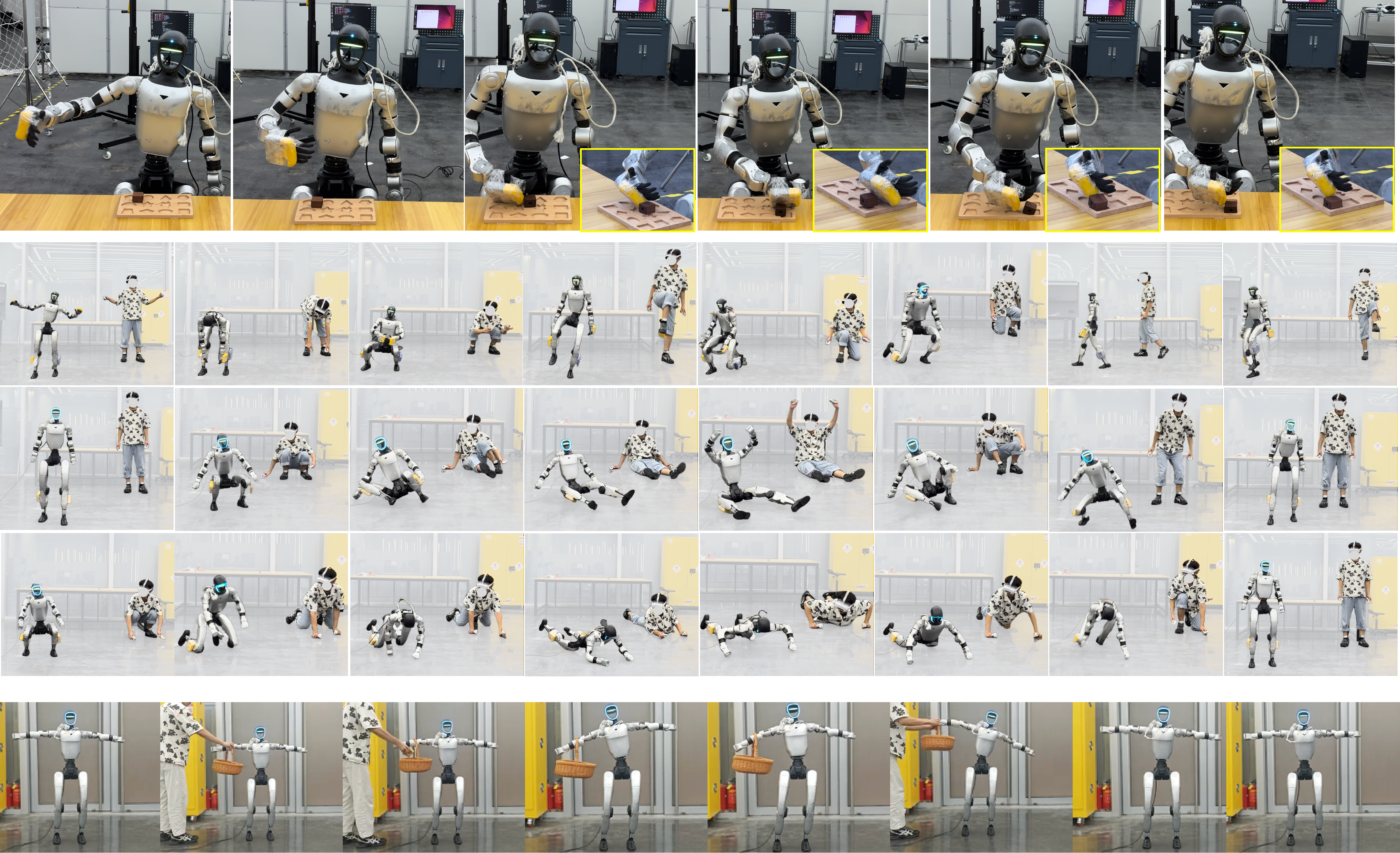}
    \caption{\textbf{Real-world in-context adaptation with ECHO.}
\textbf{Top: manipulation.} The robot performs object manipulation
with an added wrist payload; insets highlight the hand--object
interactions.
\textbf{Middle three rows: in-context whole-body behaviors.}
Under payloads attached to its limbs, the robot executes diverse
behaviors, including single-leg balance, squatting, sitting,
and transitions between standing and prone postures.
\textbf{Bottom: in-context adaptation process.}
Successive frames show the robot's postural responses as a payload
is attached to and removed from its arm.
ECHO uses interaction-derived context to guide dynamics-specific
control corrections without test-time parameter updates.
Frames within each row progress from left to right.}
    \label{fig:echo-in-context-show}
\end{figure}

\paragraph{Broad capability versus contextual adaptation.}
Figure~\ref{fig:echo-pre-show} demonstrates that ECHO-PRE supports a broad repertoire of whole-body behaviors with a single pretrained policy, ranging from ground and terrain interactions to highly dynamic aerial motions. The robot seamlessly handles substantial variations in posture, support conditions, and motion dynamics, including floor-level transitions, platform traversal, and aerial flips, highlighting the breadth and reusability of the learned behavioral capabilities.

Figure~\ref{fig:echo-in-context-show} further demonstrates that ECHO can adapt these capabilities to changing physical contexts through in-context interaction. Under additional wrist and limb payloads, the robot maintains precise manipulation and diverse whole-body behaviors, including balance, squatting, sitting, and posture transitions. Crucially, when the payload is attached or removed, ECHO adjusts its control response based on the resulting interaction history without updating model parameters at test time. This shows that ECHO complements broad pretrained capabilities with context-dependent dynamics adaptation, enabling precise behavior under previously unseen physical conditions.

\begin{table}[t]
  \centering
  \caption{\textbf{Real-robot evaluation of whole-body motion tracking.}
  We report MPJPE-L (mm, $\downarrow$; mean $\pm$ sample standard
  deviation) on four representative motions, with each checkpoint
  evaluated over 10 trials per motion. All numerical entries
  correspond to 10/10 successful trials (100\% SR), whereas
  \textit{Fail} denotes 0/10 successful trials (0\% SR).
  The two groups evaluate motion tracking without additional
  payload (top) and with a 2\,kg payload attached to each lower
  leg (bottom). Bold and underlined values indicate the best
  and second-best mean tracking errors for each motion within
  each setting, respectively.}
  \label{tab:real_mpjpe}

  \begingroup
  \footnotesize
  \setlength{\tabcolsep}{3pt}
  \renewcommand{\arraystretch}{1.18}
  \newcommand{\result}[2]{#1{\scriptsize\,\(\pm\)\,#2}}

  \begin{tabular*}{\linewidth}{@{\extracolsep{\fill}}lcccc@{}}
    \hline
    Method
    & Charleston
    & \shortstack{Single-leg\\balance}
    & \shortstack{Sit down /\\stand up}
    & \shortstack{Prone crawl /\\stand up} \\
    \hline

    \multicolumn{5}{c}{\textbf{No additional payload}} \\
    \hline

    HEFT
    & \result{51.47}{3.68}
    & \result{42.53}{2.66}
    & \textit{Fail}
    & \textit{Fail} \\

    SONIC v1.1
    & \result{54.81}{4.93}
    & \result{45.21}{3.61}
    & \textit{Fail}
    & \result{65.84}{3.70} \\

    ECHO-PRE
    & \result{\underline{48.50}}{3.21}
    & \result{\underline{42.13}}{2.45}
    & \result{\underline{53.70}}{4.79}
    & \result{\underline{55.74}}{4.39} \\

    \hline

    \textbf{ECHO-N}
    & \result{\textbf{47.98}}{3.42}
    & \result{\textbf{39.69}}{2.20}
    & \result{\textbf{49.73}}{3.90}
    & \result{\textbf{53.65}}{3.49} \\

    \hline
    \multicolumn{5}{c}{\textbf{Added payload: 2\,kg per lower leg}} \\
    \hline

    Vanilla
    & \result{64.23}{3.01}
    & \result{\underline{43.52}}{1.42}
    & \result{\underline{51.51}}{3.64}
    & \result{\underline{69.52}}{2.35} \\

    RMA-Student
    & \result{\underline{57.18}}{3.56}
    & \result{45.39}{2.51}
    & \result{75.58}{3.04}
    & \result{77.74}{2.75} \\

    \hline

    \textbf{ECHO-H}
    & \result{\textbf{52.07}}{2.21}
    & \result{\textbf{41.79}}{2.13}
    & \result{\textbf{49.95}}{3.55}
    & \result{\textbf{54.41}}{2.57} \\

    \hline
  \end{tabular*}
  \endgroup
\end{table}
\section{Limitations}
\label{sec:limitations}

Despite its broad whole-body capabilities and effective in-context
dynamics adaptation, ECHO has several limitations.
First, context learning relies on simulated rollouts and privileged
supervision, and generalization to physical effects beyond the
training randomization requires further evaluation.
Second, our memory mechanism assumes fixed physical parameters
within each session and clears the history at session boundaries.
Adaptation to unannounced changes within a session, where accumulated
context may become outdated, remains insufficiently explored.
Finally, our evaluation focuses on the Unitree G1 and
reference-conditioned whole-body control; transfer across embodiments
and integration with autonomous task planning remain untested.
We leave broader physical generalization, change-aware context
memory, and extensions across embodiments and tasks to future work.

\begin{figure}[htbp]
    \centering
    \includegraphics[width=1.0\linewidth]{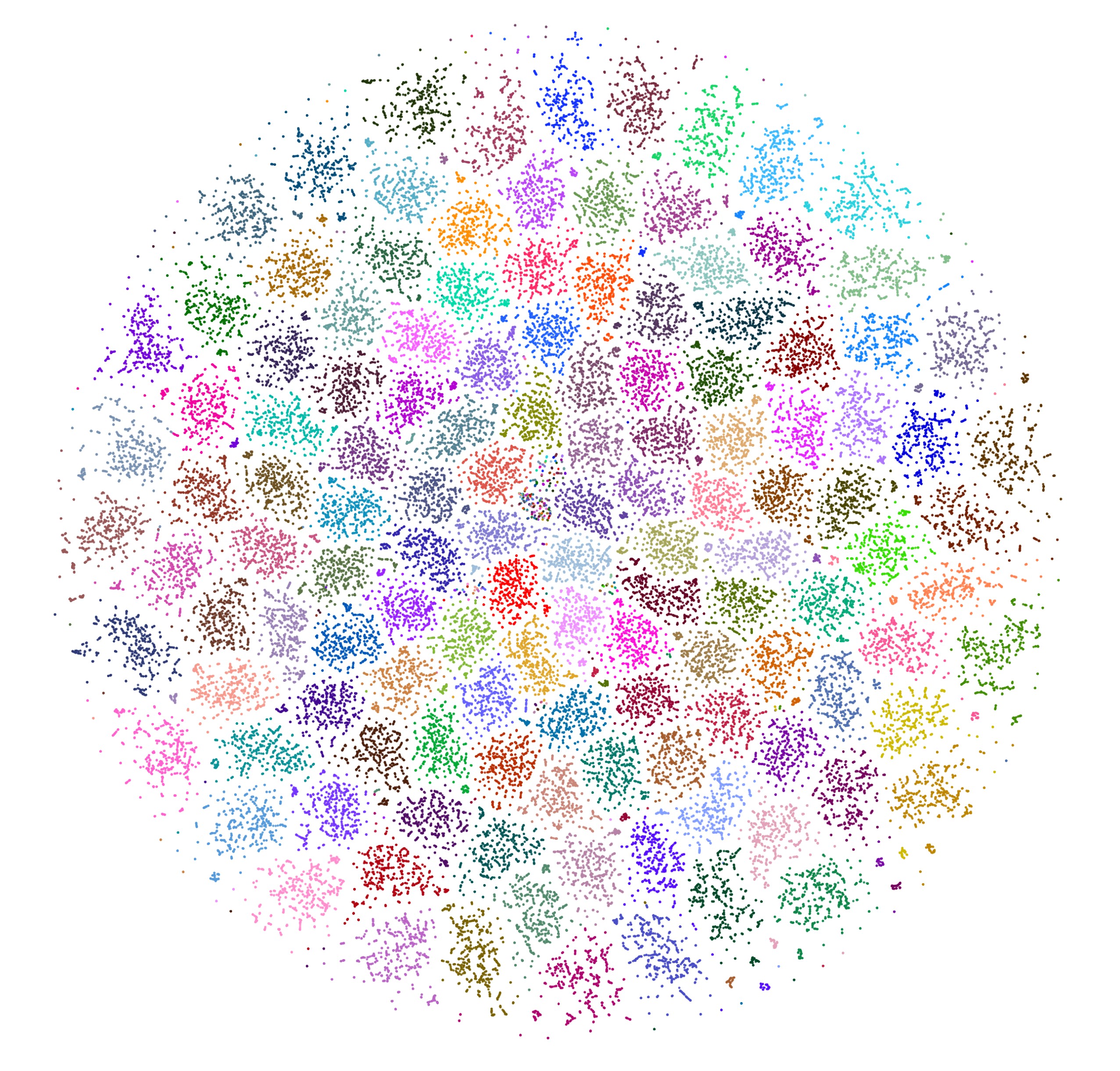}
    \caption{\textbf{Context representations on unseen AMASS motions.}
t-SNE visualization of ECHO's context embeddings from 101,967 valid
history windows collected across 100 AMASS motions and 128 physical
environments. Each point represents an individual history-window
embedding, without motion-level averaging, and each color denotes
a fixed randomized physical configuration. The environment-wise
grouping provides qualitative evidence that the learned
representations distinguish physical conditions consistently
across unseen motions.}
    \label{fig:amass100_tsne_128env}
\end{figure}

\end{document}